\documentclass{egpubl}
 
\JournalSubmission    
\usepackage[T1]{fontenc}
\usepackage{dfadobe}  

\usepackage{cite}  
\BibtexOrBiblatex
\electronicVersion
\PrintedOrElectronic
\ifpdf \usepackage[pdftex]{graphicx} \pdfcompresslevel=9
\else \usepackage[dvips]{graphicx} \fi

\usepackage{egweblnk} 

\usepackage{subfiles}

\usepackage[table,xcdraw]{xcolor}

\DeclareOldFontCommand{\rm}{\normalfont\rmfamily}{\mathrm}

\graphicspath{{figs/}{figure/}{pictures/}{images/}{./}} 

\usepackage{tabu}                      
\usepackage{booktabs}                  
\usepackage{lipsum}                    
\usepackage{mwe}                       

\usepackage{mathptmx}                  
\DeclareMathAlphabet{\mathcal}{OMS}{cmsy}{m}{n}
\DeclareSymbolFont{largesymbols}{OMX}{cmex}{m}{n}
\DeclareSymbolFont{CMlargesymbols}{OMX}{cmex}{m}{n}
\DeclareMathSymbol{\sum}{\mathop}{CMlargesymbols}{"50}

\usepackage{amsfonts}
\usepackage{color}
\usepackage{amsmath}
\usepackage{float}
\usepackage{graphicx}
\usepackage{algorithm}
\usepackage{algpseudocode}
\usepackage{wrapfig}
\usepackage{multirow}
\usepackage{makecell}
\usepackage[thicklines]{cancel}
\usepackage{tcolorbox}

\usepackage{multicol}

\newtheorem{lemma}{Lemma}

\definecolor{livingOrg}{HTML}{000000}

\newcommand{\RE}[1]{{\color{livingOrg}#1}}

\newcommand{\firc}[1]{\cellcolor[HTML]{ffcc99} \textbf{#1}}
\newcommand{\secc}[1]{\cellcolor[HTML]{fff6b2} #1}

\definecolor{pCodeBgBlue}{HTML}{E4E4FF}
\definecolor{pCodeBgOrange}{HTML}{fff3eb}
\definecolor{trf}{rgb}{0.5, 0.5, 1}

\definecolor{rred}{HTML}{000000}
\definecolor{rpurple}{HTML}{000000}
\definecolor{rbrown}{HTML}{000000}
\definecolor{rblue}{HTML}{000000}

\title[Adaptive and Iterative Point Cloud Denoising with Score-Based Diffusion Model]%
      {Adaptive and Iterative Point Cloud Denoising with Score-Based Diffusion Model}

\author[Z. Wang, M. Li, S. Xin, C. Tu]
{\parbox{\textwidth}{\centering 
        Zhaonan Wang\orcid{0009-0002-0124-4819},
        Manyi Li\thanks{Corresponding Author: manyili@sdu.edu.cn}\orcid{0000-0002-5251-0462},
        Shiqing Xin\orcid{0000-0001-8452-8723},
        Changhe Tu\orcid{0000-0002-1231-3392} 
        }
        \\
{\parbox{\textwidth}{\centering Shandong University
       }
}
}

\begin{document}


\maketitle
\begin{abstract}
Point cloud denoising task aims to recover the clean point cloud from the scanned data coupled with different levels or patterns of noise. The recent state-of-the-art methods often train deep neural networks 
to update the point locations toward the clean point cloud, and empirically repeat the denoising process several times in order to obtain the denoised results. It is not clear how to efficiently arrange the iterative denoising processes to deal with different levels or patterns of noise. 
In this paper, we propose an adaptive and iterative point cloud denoising method based on the score-based diffusion model. For a given noisy point cloud, we first estimate the noise variation and determine an adaptive denoising schedule with appropriate step sizes, then invoke the trained network iteratively to update point clouds following the adaptive schedule. \color{rred}{To facilitate this adaptive and iterative denoising process, we design the network architecture and a two-stage sampling strategy for the network training to enable feature fusion and gradient fusion for iterative denoising. }\color{black}
Compared to the state-of-the-art point cloud denoising methods, our approach obtains clean and smooth denoised point clouds, while preserving the shape boundary and details better. Our results not only outperform the other methods both qualitatively and quantitatively, but also are preferable on the synthetic dataset with different patterns of noises, as well as the real-scanned dataset.


\end{abstract}

\section{Introduction}

Point cloud is one of the most fundamental 3D representations in a majority of applications such as robotics, autonomous driving, manufacturing, etc, due to its simple data format and wide range of data sources. However, in practice, the point clouds obtained from different types of acquisition equipment often contain different levels or patterns of noise, caused by the sensor hardware, surface reflection, depth quantization, etc.
Therefore, point cloud denoising, which aims to recover the clean and accurate shape from the raw data, is often considered necessary to preprocess the scanned data and produce high-quality point clouds for the following usage.

The main challenge of the point cloud denoising task lies in filtering out the coupled noise while preserving the original shape features, merely from the observed noisy point cloud~\cite{cazals2005estimating,alexa2001point,rakotosaona2020pointcleannet}. Early works rely on heuristic assumptions, such as surface smoothness~\cite{digne2017bilateral,zaman2017density}, local compact shape descriptors~\cite{alexa2001point,  cazals2005estimating}, etc, and accordingly design geometric operators \cite{Deschaud2010POINTCN,huang2009consolidation} or optimization algorithms~\cite{Sarkar2018StructuredLM, zeng20193d} for point cloud denoising. Recently, more and more researchers have turned to geometric deep learning methods to learn the shape priors in a data-driven manner. These works often focus on how to design a proper network architecture to capture the shape features and recover the underlying surface. The learning-based methods, such as PointCleanNet~\cite{rakotosaona2020pointcleannet}, PointFilter~\cite{zhang2020pointfilter}, GDPNet~\cite{pistilli2020learning}, PDFlow~\cite{mao2022pd}, develop and train different deep network architectures to predict the displacement vectors from the input points to the underlying surface, which have shown a superior ability to capture all kinds of shape features from various noisy point clouds.

However, it is not easy to train the networks to directly estimate the accurate point positions for any shape in one forward process, which often causes over-smoothness, outlier points, surface shrinkage, etc. Therefore, to fully exploit the shape prior learned by the networks, many existing methods~\cite{Hermosilla2019total, rakotosaona2020pointcleannet, zhang2020pointfilter, luo2021score, chen2022deep,mao2022pd} choose to empirically repeat the denoising operation several times and iteratively update the point clouds to obtain better results. But unfortunately, their networks are not trained to adapt to the iterative denoising process. In other words, for every denoising timestep, the network should be aware of the previous iterations of the point cloud to estimate the shape features and update the point positions with appropriate step sizes to achieve a more faithful point cloud. More importantly, the iterative denoising process should be determined based on the noisy point cloud itself, rather than a general setting for all kinds of shapes and different levels of the coupled noise.

In this paper, we propose an adaptive and iterative point cloud denoising method based on the score-based diffusion model to deal with varying noisy point clouds. The diffusion model~\cite{ho2020denoising} is a rising generative model based on the theory of non-equilibrium thermodynamics~\cite{sohl2015deep}, which progressively destroys the structure of data distribution and trains a network to recover the data sequence in reverse order. The following score-based diffusion model~\cite{song2021scorebased} formulates the problem from the perspective of stochastic differential equation (SDE) and enables high-quality data generation from random noise with an efficient iterative sampling process. However, 
\color{rred}{ the conventional diffusion model was originally developed for the generative tasks, which aims to generate the data samples from random Gaussian noises. 
On the other hand, the point cloud denoising task aims to recover the underlying surface from the noisy point cloud, which is often considered as the random noise coupled with the original point cloud.
Therefore, to apply the diffusion model to the point cloud denoising task, 
\RE{the iteratively recovered point cloud should be aligned with the underlying surface of the input noisy point clouds, rather than directly shifting the distribution from Gaussian to learned data as done in generative tasks.}
}\color{black}

\color{rred}{To address this issue, we introduce a factor to maintain the mean of distributions of the noisy point clouds during the forward diffusion process and the reverse sampling process of the diffusion model. The formulations of the score-based diffusion model tailored for the point cloud denoising task are presented in Section~\ref{sec:formulation}, which provides the preliminary knowledge of our approach. Our methodology, as described in Section~\ref{sec:method}, includes the training strategy with two-stage sampling for the network with feature fusion and gradient fusion modules, as well as the inference algorithm with the arrangement of the adaptive denoising schedule. The feature fusion and gradient fusion aggregate the previous information and neighbor contexts to preserve the shape details. The adaptive schedule arrangement allows us to determine proper step sizes for the iterative denoising process of each noisy point cloud individually to recover the clean shape of the underlying surface. 

}\color{black}

Our contributions are summarized as follows: 
\begin{itemize}
\item \color{rred}{We propose the adaptive and iterative point cloud denoising approach based on the score-based diffusion model. It allows us to determine the adaptive denoising schedule for each individual point cloud to iteratively and efficiently recover the clean shape.
\item We propose our network with the feature fusion and gradient fusion modules, as well as the training strategy with two-stage sampling for training the feature fusion module. It allows us to use the previous information and neighbor contexts to obtain a more accurate estimation of the shape and avoid the collapse of surfaces with thin structures.}\color{black}
\item We conducted extensive comparison experiments and ablation studies to validate the effectiveness and efficiency of the proposed method. Our approach outperforms the state-of-the-art methods both qualitatively and quantitatively, and exhibits a preferable generalization ability on other patterns of noises as well as a real-scanned dataset.
\end{itemize}

\begin{figure*}[t]
    \centering
    \includegraphics[width=\linewidth]{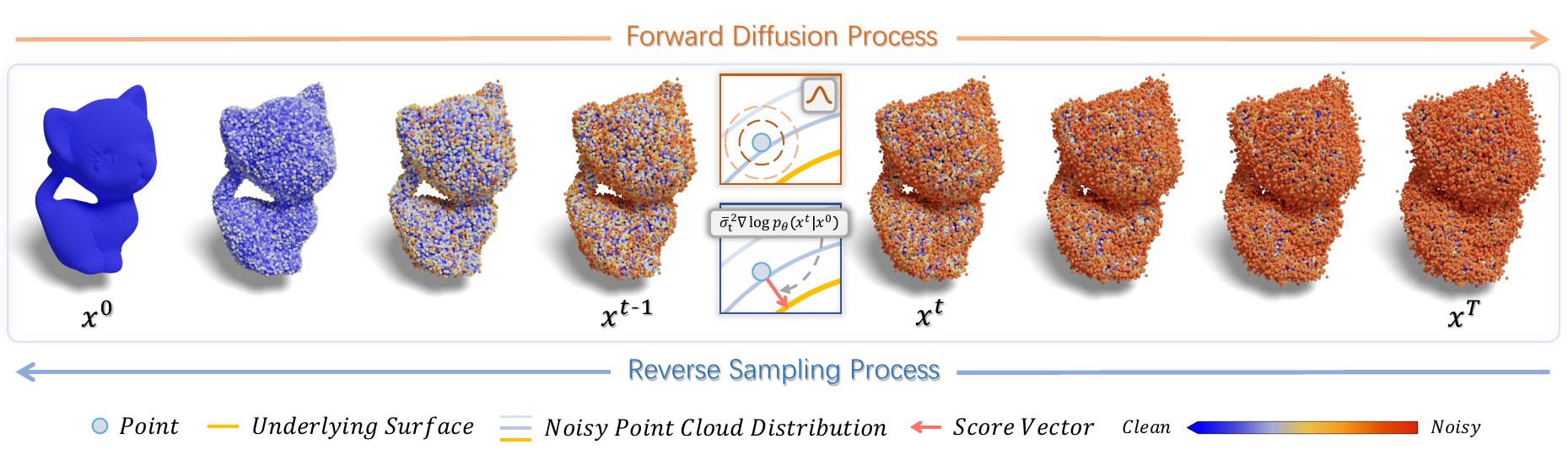}
    \vspace{-10pt}
    \caption{The \textbf{forward diffusion process} (orange arrow, from left to right) and \textbf{reverse denoising process} (blue arrow, from right to left) for 3D point clouds. The forward diffusion process progressively corrupts the clean point cloud $x^0$ with noise at each timestep following the distribution $q(x^t|x^{t-1})$, while the reverse denoising process utilizes a network at each timestep to recover the point clouds in reverse order.}
    \label{fig:diff_points}
    \vspace{-10pt}
\end{figure*}

\section{Related Work}
\subsection{Point Cloud Denoising}

There has been a large amount of literature on the point cloud denoising task \cite{cazals2005estimating, alexa2001point, rakotosaona2020pointcleannet, lipman2007parameterization, oztireli2009feature, sun2015denoising, preiner2014continuous, Deschaud2010POINTCN, Hermosilla2019total, Luo2020DifferentiableMR, zhang2020pointfilter, chen2022deep}. They rely on different shape priors to recover the underlying surface from noisy input. We classify the related methods as denoising with heuristic shape priors and deep-learning-based denoising methods.

\noindent\textbf{Heuristic Denoising Methods.} It is straightforward to assume the local smoothness of the underlying surface and define various filters to remove the high-frequency noise. For example, the bilateral filter has been extended from images~\cite{Tomasi1998BilateralFF} to 3D meshes~\cite{fleishman2003bilateral} and further point clouds~\cite{digne2017bilateral} for the denoising purpose. However, this pre-defined operator is not able to discriminate the noise and shape details of complex surfaces, especially those coupled with large-scale noise and outliers. Some methods adopt the fitting-and-projection pipeline \cite{alexa2003computing,cazals2005estimating,lipman2007parameterization}, which first approximates the underlying surface geometry with local compact descriptors \cite{alexa2001point, alexa2003computing, oztireli2009feature, cazals2005estimating} or sparse global representation~\cite{sun2015denoising, mattei2017point}, and then project the points onto the approximated surface~\cite{lipman2007parameterization, huang2009consolidation, preiner2014continuous}. Their intuition is to make a trade-off balance between the accuracy and compactness of the surface via optimization. However, they are still not good at recovering shapes with rich small-scale shape details.

Apart from the above general assumptions or objectives of all kinds of surfaces, the non-local methods~\cite{Deschaud2010POINTCN,chen2019multi} turn to patch-based priors due to the shape self-similarity. For each point, it is modified based on the other points whose neighborhood exhibits similar local patch geometry. Different surface descriptors, such as curvatures~\cite{Wang2008SimilaritybasedDO}, and polynomial coefficients of local MLS~\cite{Deschaud2010POINTCN}, are developed to measure the similarity of local patches. Recently, for a better description of the patch shape priors, Sarkar et al.~\cite{Sarkar2018StructuredLM} build a low-rank dictionary representation of the extracted patches, while GLR~\cite{zeng20193d} proposed a graph Laplacian regularization of a low-dimensional manifold model.

\noindent\textbf{Deep-learning-based methods.} Deep neural networks are good at learning the data distribution as shape priors for 3D understanding and generation tasks. Among them, the key of point cloud denoising research \cite{rakotosaona2020pointcleannet, Luo2020DifferentiableMR, luo2021score,chen2022repcd} is to develop network architectures and training strategies to learn the shape features of the underlying surface from scattered noisy point clouds.

Most of the denoising networks take the noisy point cloud as input and regress the point-wise displacements which move the input point to approach the underlying surface~\cite{rakotosaona2020pointcleannet, zhang2020pointfilter, pistilli2020learning}. These methods often require the ground-truth paired point clouds during training. It relies on the design of network architecture to extract the descriptive shape features and predict the displacement vectors. PointCleanNet~\cite{rakotosaona2020pointcleannet} is composed of two branches for outlier removal and displacement prediction respectively, which follow the design of PointNet~\cite{qi2016pointnet} and PCPNet~\cite{guerrero2018pcpnet} to be invariant to the ordering of the points. PointFilter~\cite{zhang2020pointfilter} first aligns the local patches by PCA technique and then uses an MLP network to predict the displacement vectors. However, the MLP networks are not good at capturing the shape details in different levels of context. EC-Net~\cite{yu2018ec} presents an edge-aware point cloud consolidation network for low-level noisy point clouds, with PointNet++~\cite{qi2017pointnetplusplus} architecture for multi-scale feature embedding.
GDPNet~\cite{pistilli2020learning} develops a dynamic graph convolutional network to extract the hierarchical features for point cloud denoising.
\color{rblue}{Additionally, given that the normal estimation~\cite{zhou2022refine,zhang2024norest} and point cloud denoising are often coupled together~\cite{wei2021geodualcnn,chen2023geogcn,liu2023pcdnf,yi2024pn} and both require to capture the fine geometric details, some works~\cite{wei2021geodualcnn, liu2023pcdnf} train the networks for the two tasks jointly to promote each other. The recent work  PN-Internet~\cite{yi2024pn} trains two graph convolutional networks~\cite{wei2023agconv} to refine the two tasks interactively by taking advantage of the geometric dependency between them. PathNet~\cite{wei2024pathnet} introduces a path-selective point cloud denoising paradigm that dynamically selects the most appropriate denoising path for each point and demonstrates its potential performance.}\color{black}


Although the input noisy points and ground-truth clean point cloud are assumed to be sampled on the same surface with and without noise distortion, there is not guaranteed to be a one-to-one mapping between the corresponding sampled points. In other words, the clean point cloud is just one of many implementations of the underlying surface~\cite{Hermosilla2019total}. To address this issue, some methods train the network to estimate the surface itself, instead of the displacement vectors. TotalDenoising~\cite{Hermosilla2019total} maps the point cloud to itself in combination with a spatial locality and a bilateral appearance prior, thus enabling unsupervised learning without needing access to clean examples. DMR~\cite{Luo2020DifferentiableMR} adopts the encoder-decoder architecture, where the encoder downsamples the input noisy point cloud to obtain the shape features and the decoder upsamples to reconstruct the shape as the denoised point cloud. PDFlow~\cite{mao2022pd} utilizes the normalizing flow model and disentangles the noise from the latent point representations. \color{rblue}{On the other hand, some methods seek to carefully conduct an iterative denoising process to recover the clean point clouds. The score-based denoising method~\cite{luo2021score, chen2022deep} trains the network to estimate the gradients of the points to the underlying surface and update their locations along the gradients with a heuristic and fixed schedule. They achieve better quantitative performance by iteratively updating the point locations, but the denoised point clouds still lose some small-scale shape details and exhibit a certain amount of outlier points. IterativePFN~\cite{de2023iterativepfn} incorporates the iteration in both the training and inference, by proposing an iterative point cloud filtering network to model the iteration process internally. Our work follows this line of research with the carefully designed iterative denoising process. Compared to these works, especially the closest work~\cite{luo2021score}, we arrange the adaptive denoising schedule for each individual point cloud, and utilize the previous information and neighbor contexts with our feature fusion and gradient fusion modules of the network for a better shape estimation. }\color{black}

\subsection{Diffusion Probabilistic Models}
Recently, diffusion models have emerged as a class of deep generative models, and are widely adopted in various applications, such as generation, inpainting, and super-resolution of images~\cite{ramesh2022hierarchical,rombach2022high,Fei2023GenerativeDP}. Inspired by the nonequilibrium thermodynamics theory~\cite{sohl2015deep}, the diffusion model progressively destroys the structure of the data distribution via a forward diffusion process and then trains a network to iteratively restore the data distribution in the reverse sampling process. Ho et al.~\cite{ho2020denoising} developed the Denoising Diffusion Probabilistic Model (DDPM) which is able to generate high-quality image samples. The following Denoising Diffusion Implicit Model (DDIM)~\cite{song2020denoising} and the score-based diffusion model~\cite{song2021scorebased} further improve the speed of sampling with a more flexible generation process.

There has been some research using diffusion models on image restoration tasks~\cite{kawar2022denoising, Fei2023GenerativeDP, Hussein2022ADIRAD}. They mainly focus on how to utilize the generalizable image priors of diffusion models pre-trained on large-scale datasets to solve the image restoration problem. Kawar et al.~\cite{kawar2022denoising} propose the method of Denoising Diffusion Restoration Models (DDRM) to solve any linear inverse problem with variational inference. Subsequently, Fei et al.~\cite{Fei2023GenerativeDP} introduce Generative Diffusion Prior (GDP) to model the posterior
distributions in an unsupervised sampling manner, while taking the degraded image as guidance. And ADIR~\cite{Hussein2022ADIRAD} queries K nearest neighbor images as well as the initial degraded image as guidance of the diffusion model. These works are relevant to ours in terms of iteratively updating the corrupted observations to restore clean and high-quality data following the theory of diffusion models. However, instead of taking advantage of pre-trained models, we mainly focus on how to train a diffusion model from scratch for the point cloud denoising problem, since the irregular and scattered 3D point clouds lie in a completely different data domain and the noise are deviations in all coordinates~\cite{Hermosilla2019total}.

On the other hand, some related works utilize diffusion models for point cloud tasks, such as shape generation~\cite{luo2021diffusion, Zeng2022LIONLP}, shape completion~\cite{Lyu2021ACP}, and single-view reconstruction~\cite{MelasKyriazi2023PC2PP}. They mainly adopt different network architectures for geometric deep learning to parameterize the reverse sampling process of the diffusion model. Especially with the latent diffusion model~\cite{rombach2022high}, the networks are able to learn the distribution of latent representations and generate 3D shapes 
from randomly sampled noises. In contrast to the content generation problems, our task is to recover the clean and accurate shape while respecting the observed point clouds coupled with different levels of noise. The key is to determine the adaptive denoising schedule w.r.t. the noise variance while preserving the original shape details.

\section{Score-based Diffusion for Point Cloud Denoising}\label{sec:form}
\label{sec:formulation}

The score-based diffusion model for the point cloud denoising task contains a forward diffusion process and a reverse sampling process, as in Figure~\ref{fig:diff_points}. The former defines a Markov chain of diffusion steps to progressively corrupt the original clean data with random Gaussian noise, while the latter learns to reverse the diffusion process by training a network to estimate the noise at each timestep. Once trained, given a noisy point cloud, the network is used to reduce the noise iteratively and finally restore the clean point cloud.

\color{rred}{It is worth noting that the conventional diffusion model is developed for the generative tasks, which aim to generate the data samples from random Gaussian noise. However, in our problem, the point cloud denoising task aims to recover the underlying surface from the noisy point cloud, which is often considered as the random noise coupled with the original point cloud. The noisy point clouds should maintain the mean of the distribution during the diffusion process and reverse sampling process. Otherwise, it would cause shifting positions of point clouds during the iterative process, which, when taken as the network's input, would heavily affect the network's ability to estimate the shape of the underlying surface.}\color{black}

\color{rpurple}{In this section, we present the formulation of the score-based diffusion model for the point cloud denoising task. The details of the derivations are provided in the appendix.}\color{black}


\subsection{Score-based Diffusion with Scaling Elimination}
\label{sec:diffusion}

\noindent\textbf{Forward Diffusion Process.} Starting from the clean point cloud $x^0$, the diffusion process aims to parameterize the distribution of noisy point cloud $x^t$ at timestep $t$, i.e. $q(x^t|x^0)$. According to DDPM~\cite{ho2020denoising}, it defines a Markov chain where a small amount of noise is injected at each timestep following $x^t=\sqrt{1-\beta_t} x^{t-1}+\sqrt{\beta_t} z^t, z^t\sim \mathcal N({\rm 0},{\rm I})$, with $\{\beta_t\}^{T}_{t=1}$ being a linear variance schedule to arrange the noise injection during the diffusion process. The distribution is then derived as $q(x^t|x^0)=\mathcal{N}(x^t;\sqrt{\bar \alpha_t}x^0,(1-\bar\alpha_t){\rm I})$, with $\alpha_t=1-\beta_t$ and $\bar\alpha_t=\prod^t_{s=1}\alpha_s$. Therefore, we can sample the noisy point cloud as $x^t=\sqrt{\bar \alpha_t}x^0+\sqrt{1-\bar\alpha_t}z,z\sim \mathcal N({\rm 0},{\rm I})$. 

However, \color{rred}{the above formulation in the conventional diffusion model requires scaling the original point cloud $x^0$ with $\sqrt{\bar \alpha_t}$, which are inconsistent for the noisy point clouds at different timesteps. The shifting among noisy point clouds, i.e. the input points of the noise prediction network, would heavily affect the network's ability in the reverse process. To maintain the distribution of point clouds, }\color{black} we should rewrite the sampling of noisy point clouds in the form of $x^t=x^0+\bar\sigma_t z$, where $\bar\sigma_t$ represents the total noise injected to the original point cloud until the timestep $t$. So we parameterize the distribution of noisy point clouds as
\begin{equation}\label{eq:diffusion_equation}
    q(x^t|x^0)=\mathcal N(x^t;x^0,\ (1-\bar\alpha_t)/\bar\alpha_t{\rm I}).
\end{equation}
We also need to reformulate the Markov chain to match the noisy point cloud distribution (Eq~\ref{eq:diffusion_equation}). The joint distribution becomes
\begin{equation}
\begin{aligned}\label{eq:perstep_diffusion}
q(x^{1:{T}}|x^0)=\prod^{T}_{t=1}q(x^t|x^{t-1}), q(x^t|x^{t-1})=\mathcal N(x^t;x^{t-1},\ \beta_t/\bar\alpha_t{\rm I}).
\end{aligned}
\end{equation}

In this way, the above formulation avoids the shifting issue by eliminating the scaling factor $\sqrt{\bar \alpha_t}$. In addition, it provides the correspondence between the linear schedule $\{\beta_t\}^{T}_{t=1}$ and the noise variance $\bar\sigma_t$ of the noisy point cloud $x^t$. That is, at timestep $t$, the current point cloud $x^{t-1}$ is corrupted with noise variance $\sigma_t^2=\beta_t/\bar\alpha_t$ to obtain $x^t$. Compared to the original clean point cloud $x^0$, we have the total noise variance 
\begin{equation}\label{eq:noise_variance}
\bar\sigma_t^2=(1-\bar\alpha_t)/\bar\alpha_t, \quad \bar\alpha_t=\prod^t_{s=1}\alpha_s, \quad
\alpha_t=1-\beta_t.
\end{equation}
This correspondence enables us to determine an adaptive and iterative denoising schedule $\{\beta_t\}^{T}_{t=1}$ by estimating the noise variance of a given noisy point cloud, as described in Section~\ref{sec:algorithm}.

\noindent\textbf{Reverse Sampling Process.} The sampling process aims to recover point cloud sequence $x^{0:T}=\{x^0,\dots,x^T\}$ in a reverse order, where $x^0$ represents the clean point cloud lying on the underlying surface.
The reverse process is also parameterized as a Markov chain
\begin{equation}
\begin{aligned}\label{dif:pThetaDefine}
    p_\theta(x^{0:T})=p(x^T)\prod^T_{t=1}p_\theta(x^{t-1}|x^{t}),\\
    \color{rbrown}{p_\theta(x^{t-\Delta}|x^{t})=\mathcal N(x^{t-\Delta};\mu_\theta(x^{t},\Delta),\Sigma),}\color{black}
\end{aligned}
\end{equation}
where $\mu_\theta$ is estimated by a denoising network with parameters $\theta$ and the variance $\Sigma$ can be computed in closed form w.r.t. the arranged schedule $\{\beta_t\}^T_{t=1}$. \color{rbrown}{The above formulation with $\Delta\in [1,t]$ allows us to estimate the noisy point cloud at any previous timestep before $t$. }\color{black} Based on Bayes' theorem and parameterization, we have 
\begin{equation}\label{eq:mu_theta}
\begin{aligned}
    \color{rbrown}{
    \mu_\theta=x^{t}+(\sqrt{\frac{1-\bar\alpha_{t-\Delta}(1+\Sigma_\eta)}{\bar\alpha_{t-\Delta}}}-\sqrt{\frac{1-\bar\alpha_{t}}{\bar\alpha_{t}}})z_\theta(x^{t}),
    }\color{black}
\end{aligned}
\end{equation}
where $z_\theta(x^t)$ is the estimated noise from the network at timestep $t$ and $\Sigma_\eta=\eta\Sigma$ is controlled by the variable $\eta\in [0,1]$.

\begin{figure*}[ht]
    \centering
    \includegraphics[width=\linewidth]{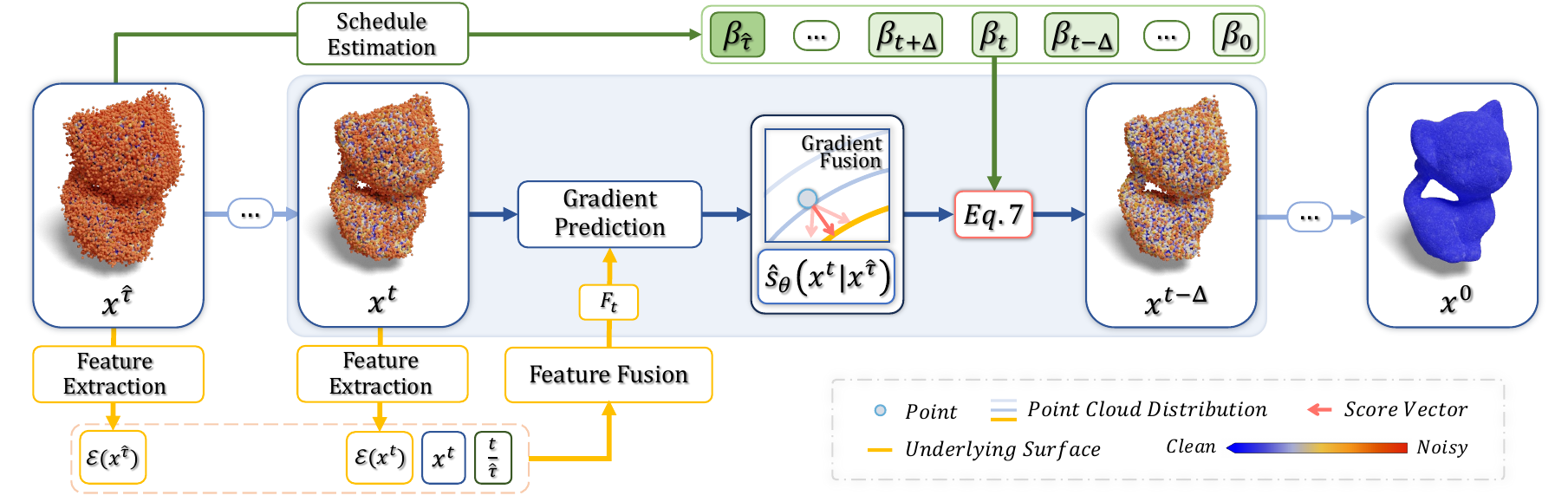}
    \vspace{-10pt}
    \caption{\textbf{The iterative denoising pipeline of the proposed method.} Given a noisy point cloud, our method first estimates its adaptive schedule $\{\beta_t\}^{\hat\tau}$ which controls the amount of noise removed at each timestep. During the iterative denoising process, at each timestep, we estimate the shape feature $\mathcal{E}(x^t)$ and predict the score vectors \RE{$\hat{s}_\theta(x^t|x^{\hat\tau})$} at the point locations. The score vectors are used to update the point cloud with Equation~\ref{eq:sampling} following the adaptive schedule until obtaining the denoised point cloud $x^0$. Note that we use feature fusion and gradient fusion while invoking the trained networks to preserve the detailed shape features.}
    \label{fig:pipeline}
    \vspace{-20pt}
\end{figure*}

We further define the score vector $s_\theta(x^t)$ as gradient of its log-likelihood function $-\bar{\sigma}_t z_\theta\propto 
\bar{\sigma}^2_t\nabla_x\log{[p_{\theta}(x^t|x^0)]}=s_\theta(x^t)$ \RE{, pointing from $x^t$ to underlying surface}. Then the sampling process can be expressed as
\begin{equation}\label{eq:sampling_org}
\color{rbrown}{
    x^{t-\Delta}=\underbrace{x^t+(1-\sqrt{\frac {(1-\bar\alpha_{t-\Delta}(1+\Sigma_\eta))\bar\alpha_t}{(1-\bar\alpha_t)\bar\alpha_{t-\Delta}}})s_\theta(x^t)}_{\mu_\theta}
    +\underbrace{z\sim\mathcal N(0,\Sigma_\eta{\rm I})}_{random\ noise}.
}\color{black}
\end{equation}
Note that the reverse sampling process becomes the generative DDPM~\cite{ho2020denoising}
with
\color{rbrown}{$\Sigma={\bar\sigma^2_{t-\Delta}(\bar\alpha_{t-\Delta}-\bar\alpha_{t})}/{((1-\bar\alpha_{t})\bar\alpha_{t-\Delta})}$ }\color{black}
and $\eta=1$, where the random noise sampling may blur the detailed shape feature. Instead, we perform deterministic sampling by setting $\eta=0$. The reverse sampling process (Eq~\ref{eq:sampling_org}) finally becomes
\begin{equation}\label{eq:sampling}
\color{rbrown}{
    x^{t-\Delta}=x^t+(1-\sqrt{\frac{(1-\bar{\alpha}_{t-\Delta})\bar{\alpha}_{t}}{(1-\bar{\alpha}_{t})\bar{\alpha}_{t-\Delta}}})s_\theta(x^t).
}\color{black}
\end{equation}
Therefore, we only need to train a network to predict the score vectors, i.e. the gradients towards the underlying surface, and iteratively update the point locations based on Eq~\ref{eq:sampling}. The advantages brought by the score-based diffusion model are that the network can predict better score vectors at the low-density regions to alleviate outliers in the denoised point clouds and it allows us to flexibly arrange the iterative denoising schedule during sampling process.

\subsection{Training Objective}
\label{sec:training_objective}

Intuitively, the training is to minimize the \color{rpurple}negative \color{black} log-likelihood of the denoised point cloud. We define the trainable objective $\mathcal L$ as:
\begin{equation}
    \begin{aligned}
        \min_\theta &\mathbb{E}_{q(x^0)}[-\log{p_\theta(x^0)}]\\
        &\le\min_\theta \mathbb E_q\bigg[\color{rbrown}{\sum_{t> 1,\Delta}
        D_{KL}({q(x^{t-\Delta}|x^t,x^0)} \ ||\ {p_\theta(x^{t-\Delta}|x^{t})})}\color{black}
        \bigg]\\
        &\ \ \ \ \ \ \ \ \ \ \ \ \ \ \ \ +{D_{KL}({q(x^{T}|x^0)}\ ||\ {p(x^{T})})}-{\log {p_\theta(x^{0}|x^{1})}}\\
        &=:\mathcal L(x^{0:{T}},\{\beta_t\}^{T}_{t=1}).
    \end{aligned}
\end{equation}

To make the training simpler and more efficient, following \cite{ho2020denoising}, we randomly choose the diffusion timesteps in the loss function. In addition, we define a weighting term related to $\bar\sigma_t$, i.e. the total scale of noise coupled with the clean point cloud, to encourage the network to focus more on the denoising performance near the underlying surface. In other words, the weighting term has a larger value when $t$ is small. Our loss function is
\begin{equation}\label{dif:lossSimp}
    \mathcal L=\mathbb E_{q,t>0}\bigg[\bigg|\bigg|\Big(\frac{1-\lambda}{\bar{\sigma}_t}+\lambda\Big)\Big(s_\theta(x^t)-S(x^t)\Big)\bigg|\bigg|^2_2\bigg],
\end{equation}
where $\lambda$ is the parameter to control the weighting term, \color{rpurple}{$s_\theta(x^t)$ is the network-estimated score vector while $S(x^t)$ is the ground truth score vector. }\color{black} We use $\lambda=0.99$ in all the experiments.

\section{Method}
\label{sec:method}

Our approach includes training the network with feature fusion and gradient fusion modules, as well as adaptive and iterative denoising during inference. In this section, we first introduce the network architecture and the details of each network module in Secion~\ref{sec:network}, then the training algorithm in Section~\ref{sec:training_algorithm}, and the inference algorithm in Section~\ref{sec:algorithm}. Figure~\ref{fig:pipeline} illustrates the iterative denoising process during inference with the trained network.

\subsection{Network Architecture}
\label{sec:network}

\color{rpurple}{Our denoising network takes a noisy point cloud $x^t=\{x^t_i\}^n_{i=1}$ as input, and predicts the score vector for each of the points, i.e. $\hat{s}_\theta(x^t)$. The network is sequentially composed of a feature extraction module $\mathcal{E}$ to extract the point-wise features, a feature fusion module $\mathcal{F}$ to integrate the point features of the current point cloud and those of the original noisy point cloud, a gradient prediction module $\mathcal{G}$ to predict the points' gradient vectors based on their correlated point features, and a gradient fusion module $\xi$ to fuse the multiple versions of the predicted gradients (based on the features of different correlated points) into the predicted score vectors.}\color{black}

The \emph{feature extraction module $\mathcal{E}$} is implemented as a dynamic graph convolutional neural network (DGCNN)~\cite{wang2019dynamic} to deal with the point clouds with various shapes. The network contains a stack of dynamic graph convolutional layers, which search the neighbors of each point in terms of feature similarity and aggregate the context information to obtain the multi-scale point features. We use $\mathcal{E}(x^t)$ to denote the extracted point features for point cloud $x^t$.

\color{rbrown}{
The \emph{feature fusion module $\mathcal{F}$}\RE{, designed to blend the coarse shape information of $x^T$ with the detail feature of the recovered $x^t$,} fuses the point features of the current iteration $\mathcal{E}(x^t)$ and those of the input point cloud $\mathcal{E}(x^T)$ \RE{in a way that leverages the timestep and current position}:
\begin{equation}
    F_t={\mathcal{F}}(x^t, \frac{t}{T}, \mathcal{E}(x^t), \mathcal{E}(x^T)).
\end{equation}
Specifically, we first fuse the positional encoding of $x^t$ and the embedding of relative timestep $\frac{t}{T}$ as $e$. The feature $e$ is then fused with $\mathcal{E}(x^t)$ to obtain \RE{the weight vector} $e^t$, and with $\mathcal{E}(x^T)$ to obtain $e^T$. We finally fuse all of them to form the fused feature $F_t=MLP(\mathcal{E}(x^t)\cdot e^t + \mathcal{E}(x^T)\cdot e^T)$, with $\cdot$ represents element-wise multiplication. All the above fusions are conducted with corresponding 3-layer MLP networks.
}\color{black}


\color{rbrown}The \emph{gradient prediction module $\mathcal{G}$} predicts the gradient vector of an arbitrary point $v$ supported by a correlated point-wise feature $F_{t,i}$ within the local field of $v$, and the importance weight indicating the effectiveness of the $F_{t,i}$ at $v$. This module is implemented as an MLP network composed of four residual blocks. Taking the arbitrary point $v$ and the point-wise feature $F_{t,i}$ with its corresponding point $x^t_i$ as input, we first transform $v$ to the relative coordinate for normalization purpose, i.e. $v-x^t_i$, and input it to the network:
\begin{equation}
g_{v,i},w_{v,i}=\mathcal{G}(v-x^t_i, F_{t,i}),
\end{equation}
where $g_{v,i}$ and $w_{v,i}$ denote the predicted gradient vector and the importance weight, respectively. 
\color{rpurple}During both the training and testing, we sample $v$ from the noisy point cloud $x^t$, and query the neighbor point $x^t_i$ from the same point cloud.\color{rbrown}

Considering that a point $v$ involves many different points within its local neighborhood, we query $k=32$ nearest neighbors as different $x^t_i$, thus producing multiple versions of the gradient at the location $v$ based on different point-wise features $F_{t,i}$. We further use a gradient fusion module to obtain a reasonable and accurate gradient estimation. A simple average fusion sometimes causes collapse, especially when the underlying surface contains thin structures. We define a weighted gradient fusion function based on the important weights predicted by the gradient prediction module.

Our \emph{gradient fusion function} $\xi$ is defined as a weighted function:
\begin{equation}
    \RE{\xi(v|x^t,x^T)}=\sum^k_{i=1}g_{v,i}\cdot {\rm SoftMax}(\{w_{v,i}\})_i.
    \label{eq:gradientFusion}
\end{equation}
where ${\rm SoftMax}(\cdot)$ filters the uncertain gradient vectors indicated by important weights. $\cdot$ represents element-wise multiplication.
It prevents the thin structures from collapsing into degenerated surfaces.
\color{black}
\RE{Finally, we can obtain the predicted score vector for each of the points
\begin{equation}
    \hat{s}_\theta(x^t|x^T)=\{\xi(v|x^t,x^T)|v\in x^t\}^n.
\end{equation}
}

\subsection{Training Algorithm}
\label{sec:training_algorithm}

    
\begin{algorithm}[tbp]
\caption{Training Algorithm}
\label{alg:training}
\textbf{Input:} The ground-truth clean point cloud $\mathcal{X}$
; the training schedule $\{\beta_t\}^T_{t=1}$; the length of the training schedule $T$. \\
\textbf{Initialize:} The feature extraction module $\mathcal{E}$; the feature fusion module $\mathcal{F}$; the gradient estimation module $\mathcal{G}$.
\begin{algorithmic}[1]
\Repeat
    \Statex \ \ \ -------------------- Patch Initialization -----------------------------
    \State Sample a random point $p\in \mathcal{X}$
    \State Initialize the patch $x^0\gets kNN(p,\mathcal{X})$ centered at $p$
    \State Initialize patch mask $M$ to highlight $K_p$ nearest points of $p$ 
    \Statex \ \ \ ------------- Stage 1: sampling and process $x^t$ ------------------
    \State $t \sim \text{Uniform}(\{t_{min}, \ldots, T\})$, $\epsilon \sim \mathcal{N}(0, \bar{\sigma}_{t}^2 {\rm I})$
    \State Construct a noisy point cloud $x^{t} \gets x^0 + \epsilon$
    

    \State Extract the fused feature $F_{t}\gets \mathcal{F}(x^{t},\frac{t}{t},\mathcal{E}(x^{t}),\mathcal{E}(x^{t}))$
    \State Predict the score vectors \RE{$\hat{s}_\theta(x^{t}|x^t)$} \Comment{Eq.~\ref{eq:gradientFusion}}
    \Statex \ \ \  ------------- Stage 2: sampling and process $x^{t-\Delta}$ --------------
    
    \State $\Delta \sim \text{Uniform}(\{1, \ldots, t\})$
    \State Update the noisy point cloud \Comment{Eq.~\ref{eq:sampling}} \par $x^{t-\Delta} \gets x^{t} + (1-\sqrt{\frac{(1-\bar{\alpha}_{t-\Delta})\bar{\alpha}_{t}}{(1-\bar{\alpha}_{t})\bar{\alpha}_{t-\Delta}}})({M}\RE{\hat{s}_\theta(x^{t}|x^t))}+\neg{{M}}S(x^{t}))$
    
    \State Extract fused feature $F_{t-\Delta}\gets \mathcal{F}(x^{t-\Delta},\frac{t-\Delta}{t},\mathcal{E}(x^{t-\Delta}),\mathcal{E}(x^{t}))$
    \State Predict the score vectors \RE{$\hat{s}_\theta(x^{t-\Delta}|x^t)$} \Comment{Eq.~\ref{eq:gradientFusion}}
    \Statex \ \ \ ---------------------------- Loss --------------------------------------

    \State Compute GT score vectors $S(x^{t})$ and $S(x^{t-\Delta})$ \Comment{Eq.~\ref{eq:computeGT}}
    \State Update parameters with \par $\mathcal{L}\gets
    \mathbb E\Big[\Big|\Big|\big(\frac{1-\lambda}{\bar{\sigma}_{t}}+\lambda\big){M}\big(\RE{\hat{s}_\theta(x^{t}|x^t)}-S(x^{t})\big)\Big|\Big|^2_2\Big]$\par
    $\quad\quad + \mathbb E\Big[\Big|\Big|\big(\frac{1-\lambda}{\bar{\sigma}_{t-\Delta}}+\lambda\big){M}\big(\RE{\hat{s}_\theta(x^{t-\Delta}|x^t)}-S(x^{t-\Delta})\big)\Big|\Big|^2_2\Big]$
\Until{converged}
\end{algorithmic}
\end{algorithm}

\color{rbrown}
We develop a two-stage sampling and patch-wise training algorithm for our network. On one hand, training on small patches rather than the entire point clouds improves the network's generalization ability, especially when the training set only contains a few point clouds. On the other hand, the training of the diffusion model often samples the noisy data at a random timestep of the diffusion process for each training iteration. Since our feature fusion module requires the point features of both the current and original noisy point cloud, the two-stage sampling strategy aims to obtain the two noisy point clouds at one training iteration.

Algorithm~\ref{alg:training} describes our two-stage sampling and patch-wise training algorithm. Given the ground-truth clean point cloud $\mathcal{X}$, in each iteration of the training, we conduct the patch initialization then the two-stage sampling to compute the loss. Specifically, to initialize a patch, we randomly crop a local 1000-point patch from the entire point cloud as $x^0$, and make the point-wise mask ${M}$ which highlights $K_p=256$ points within a smaller region around the patch center. 
The mask ${M}$ is used to regularize the loss, i.e. only the loss for points highlighted by the mask is used for the training, to ensure that every point has sufficient local neighborhood information for the gradient prediction and gradient fusion in this training iteration. 

After the patch initialization is the two-stage sampling. In the first stage, it samples the noisy point cloud $x^t$ at timestep $t$ and predicts the score vectors \RE{$\hat{s}_\theta(x^{t}|x^t)$}. In the second stage, it samples another noisy point cloud $x^{t-\Delta}$ by updating the positions of $x^t$ with \RE{$\hat{s}_\theta(x^{t}|x^t)$}, then predicts the corresponding score vectors \RE{$\hat{s}_\theta(x^{t-\Delta}|x^t)$}. Note that the goal of sampling point clouds $x^t$ and $x^{t-\Delta}$ is to imitate an iteration step during inference, involving both an original and a current noisy point cloud. It allows us to learn the feature fusion module of the network during training.

The loss function, i.e. 14th line in Algorithm~\ref{alg:training}, is defined as the MSE loss between the predicted score vectors \RE{$\hat{s}_\theta(x^{t}|x^t)$} (and \RE{$\hat{s}_\theta(x^{t-\Delta}|x^t)$}) and the ground-truth score vectors $S(x^t)$ (and $S(x^{t-\Delta})$). \color{black}\color{rpurple}{Following~\cite{luo2021score,chen2022deep}, we use the displacement vectors from the point locations $x^t$ to their nearest neighbor in the clean point cloud $x^0$ to approximate the ground-truth score vectors $S(x^t)$ \RE{ pointing from $x^t$ to the underlying surface}:
\begin{equation}
\label{eq:computeGT}
    {
    S(x^t)=NN(x^t,x^0)-x^t,
}
\end{equation}
where $NN(x^t,x^0)$ returns the nearest neighbor in $x^0$ for each point in $x^t$. Note that although the exact score vector can be directly computed w.r.t. the clean meshes in the dataset, the adopted approximation significantly reduces the training time per iteration and introduces negligible inaccuracy in our experiments.}\color{black}

To summarize, we only need the clean point clouds as the training set and sample the noisy patches during training to compute the scores as ground-truth. We assume the noise variance does not exceed 3\% in our experiments, i.e. $\bar\sigma_t\le 3\%$. And we set the number of diffusion steps as $T=1000$ during training in order to learn the score distribution at low-density regions. Therefore we have the linear schedule $\{\beta_t\}$ ($\beta_0=0$) with $\beta_{T}=2e^{-6}$ in order to satisfy $\bar\sigma_t\in(0,0.03]$.

\subsection{Adaptive and Iterative Denoising Algorithm}
\label{sec:algorithm} 

After training, given a noisy point cloud $\mathcal{X}$, we estimate the noise variance, determine an adaptive denoising schedule to iteratively update point locations, and finally restore the clean point cloud $\hat{\mathcal{X}}$. The detailed algorithm is presented in Algorithm~\ref{alg:sampling}. \color{rpurple}Following SBD~\cite{luo2021score} and IterPFN~\cite{de2023iterativepfn}, we use farthest point sampling to construct $R$ overlapping patches, i.e. $\{\mathcal{X}_r\}^R$.\color{black}

\noindent\textbf{Adaptive Schedule Arrangement.} We use the trained denoising network to estimate the noise variance of the input point cloud $\mathcal{X}$. The network predicts the scores \RE{$\hat{s}_\theta(\mathcal{X}|\mathcal{X})$}, whose variance provides a primary guess of the noise variance
\begin{equation}\label{eq:primarysigma}
    \bar\sigma^2={\rm Var}\big(\ \vert\vert \RE{\hat{s}_\theta(\mathcal{X}|\mathcal{X})} \vert\vert\ \big),
\end{equation}
where we directly concatenate the estimated score vectors for all the patches as \RE{$\hat{s}_\theta(\mathcal{X}|\mathcal{X})$}.

Then our adaptive schedule is to find a sub-sequence from $\{\beta_0,\dots,\beta_{T}\}$, denoted as $\{\beta_0,\dots,\beta_{\hat \tau}\}$, so that the corresponding noise variance $\bar\sigma^2_{\hat \tau}$ based on Equation~\ref{eq:noise_variance} match with $\bar\sigma^2$ estimated with Equation~\ref{eq:primarysigma}. However, there's no closed-form solution to solve $\beta_{\hat \tau}$ based on $\bar\sigma^2_{\hat \tau}$. So we use the linear search algorithm:
\begin{equation}\label{eq:optimalbeta}
    \beta_{\hat \tau}=\mathop{\rm argmin}_{\beta_{\hat \tau}\in \{\beta_t\}^T}\ \big\vert \bar\sigma^2_{\hat \tau}-\bar\sigma^2 \big\vert.
\end{equation}
Note that $\beta_{\hat \tau}$ is selected from the sequence $\{\beta_0,\dots,\beta_{T}\}$, which is used during the training. This search terminates when finding the $\hat \tau$ that best satisfies the search objective.

\color{rbrown}After obtaining $\beta_{\hat \tau}$ and the corresponding $\hat \tau$, we linearly interpolate $\hat \tau$ with $\Delta=\hat\tau/L$ and compute their corresponding schedule, i.e. an $L$-element sequence $\{\beta^*_0,\dots,\beta^*_{\hat \tau}\}$, where $L$ is the number of iteration in the following denoising algorithm. To further align the schedules used during training and inference, we replace each element with the closest one in the training schedule, thus obtaining $\{\beta_0,\dots,\beta_{\hat \tau}\}$. We empirically set the length of the adaptive denoising schedule $L=5$ for efficiency purposes. \color{black}

\begin{algorithm}[tbp]
\caption{Adaptive and Iterative Denoising Algorithm}
\label{alg:sampling}
\textbf{Input:} The noisy point cloud $\mathcal{X}$
, the training schedule $\{\beta_t\}^T$, the length of the adaptive schedule $L$; network modules $\mathcal{E}$, $\mathcal{F}$, $\mathcal{G}$ with trained weights from Algorithm~\ref{alg:training}.
\begin{algorithmic}[1]
\State Construct $R$ overlapping patches $\{\mathcal{X}_r\}^R$ using FPS
\Statex --------------- Adaptive Schedule Arrangement --------------------
\State Predict primary noise variance $\bar\sigma^2$ of input $\mathcal{X}$ \Comment{Eq.~\ref{eq:primarysigma}}
\State Estimate schedule $\{\beta_t\}^{\hat\tau}$ \Comment{Eq.~\ref{eq:optimalbeta}}
\Statex ---------------- Patch-wise Iterative Denoising ---------------------

\For{$r = 1$ to $R$}
    \State Initialize $x^{\hat\tau}\gets \mathcal{X}_r,\ \Delta=\hat\tau/L$
    \State Extract feature $\mathcal{E}(x^{\hat\tau})$ of $x^{\hat\tau}$
     \For{$l = L$ to $1$}
        \State $t\gets \text{Round}(l\cdot \Delta)$ \Comment{$t$ and $t-\Delta$ are rounded to integers.}
        \State Extract feature $\mathcal{E}(x^t)$ of $x^t$
        \State Compute fused feature $F_{t} \gets \mathcal{F}(x^t,\frac{t}{\hat\tau},\mathcal{E}(x^t),\mathcal{E}(x^{\hat\tau}))$
        \State Predict the score vectors \RE{$\hat{s}_\theta(x^{t}|x^{\hat{\tau}})$} \Comment{Eq.~\ref{eq:gradientFusion}}
        \State Update the noisy point cloud \Comment{Eq.~\ref{eq:sampling}} \par \quad $x^{t-\Delta} \gets x^t +(1-\sqrt{\frac{(1-\bar\alpha_{t-\Delta})\bar\alpha_{t}}{(1-\bar\alpha_{t})\bar\alpha_{t-\Delta}}})\RE{\hat{s}_\theta(x^{t}|x^{\hat{\tau}})}$
    \EndFor
    \State $\hat{\mathcal{X}}_r \gets \{x_i^{0}\}_{i=1}^n$
\EndFor
\State Stitch $R$ output patches $\{\hat{\mathcal{X}}_r\}^R$ as point cloud $\hat{\mathcal{X}}$\\
\textbf{Output:} The restored point cloud $\hat{\mathcal{X}}$.

\end{algorithmic}
\end{algorithm}

\noindent\textbf{Iterative Denoising Algorithm.} 
\color{rpurple}{After determining the adaptive schedule, we conduct the iterative denoising process for each patch separately, following the same adaptive schedule. 
The algorithm takes a patch of noisy point cloud $\mathcal{X}_r$ as input and outputs the restored point cloud $\hat{\mathcal{X}}_r$ for this patch. Finally, to stitch the restored patches $\{\hat{\mathcal{X}}_r\}^R$ together, we select each point's most reliable position from the involved patches. That is, if a point is covered by multiple patches, we select the patch where this point is closer to the patch center and use the corresponding restored position to form the denoised point cloud for the entire shape, i.e. $\hat{\mathcal{X}}$.}\color{black}

\section{Experiments}
\begin{figure*}
    \centering
    \includegraphics[width=\linewidth]{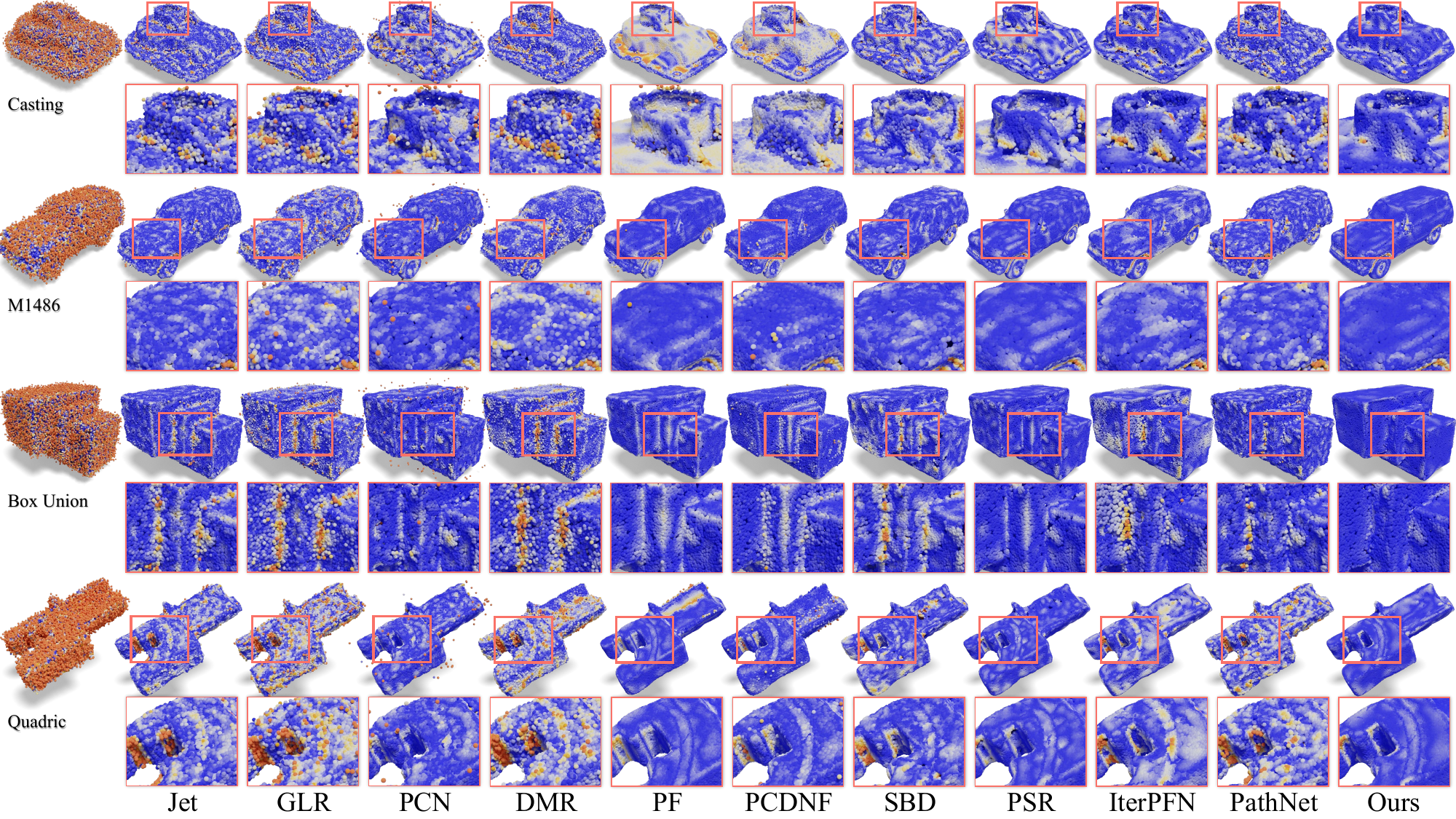}
    \vspace{-10pt}
    \caption{\textbf{Visual comparisons of denoising methods under $3\%$ Gaussian noise.} Points colored orange are farther away from the ground truth surface and blue is closer. The results are denoised from the noisy input with 50k points.}
    \label{fig:baseline}
    \vspace{-10pt}
\end{figure*}
\begin{figure*}
    \centering
    \includegraphics[width=\linewidth]{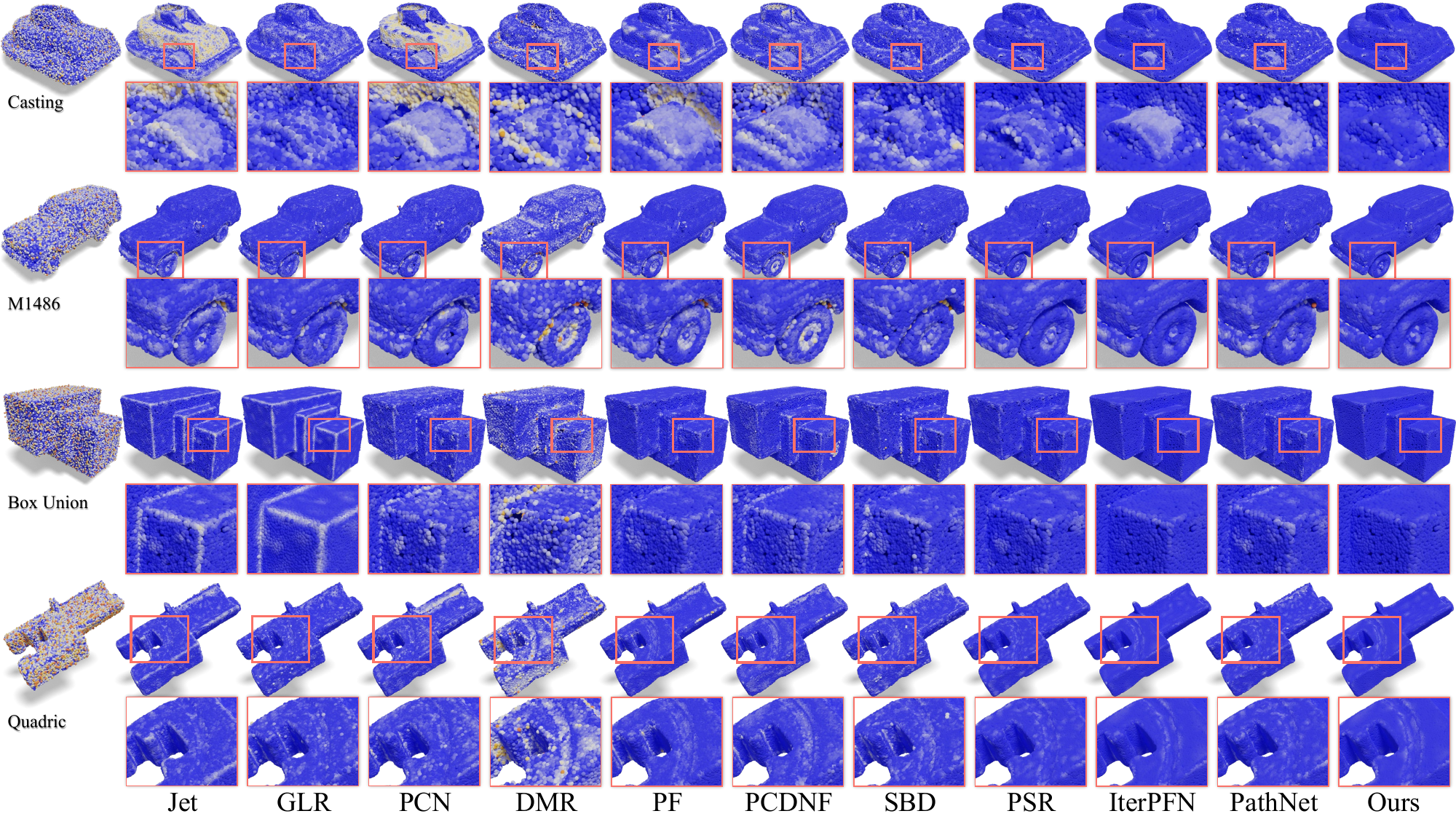}
    \vspace{-10pt}
    \caption{\textbf{Visual comparisons of denoising methods under $1\%$ Gaussian noise.} Points colored orange are farther away from the ground truth surface and blue is closer. The results are denoised from the noisy input with 50k points.}
    \label{fig:detail}
    \vspace{-10pt}
\end{figure*}

\label{sec:results}

\subsection{Setup}
\noindent\textbf{Datasets.} 
Following the recent related works \cite{yu2018pu, luo2021score}, we use the training set proposed by PU-Net \cite{zhang2020pointfilter} which includes 40 meshes. The clean point clouds are obtained with the Poisson disk sampling algorithm and are normalized into a unit sphere. Point cloud data augmentation such as random rotation and scaling is adopted during training to reduce the overfitting problem. Note that our network processes the patch-wise point clouds which enable the generalization to all kinds of shapes. 

For testing, we compare with the state-of-the-art methods on the synthetic test sets of PU-Net \cite{yu2018pu,luo2021score}, PC-Net~\cite{rakotosaona2020pointcleannet} and MOD-Net \cite{huang2022modnet}, which contain 20, 10, 10 meshes respectively. The MOD-Net test set contains meshes with rich geometric details which helps us to examine the detail-preserving ability of different methods. We remove the meshes in the MOD test set which are included in the training set and the other test sets for the quantitative evaluation. In addition, we also examine the generalization ability on the real-scanned point clouds from the Paris-rue-Madame Database~\cite{serna2014paris}, as well as the point clouds coupled with different patterns of noise.

\noindent\textbf{Metric.} 
We employ two metrics commonly adopted in previous works for the quantitative evaluation: Chamfer distance (CD)~\cite{fan2017point} and point-to-mesh distance (P2M)~\cite{ravi2020accelerating}. 

\noindent\textbf{Implementation Details.}
We implement our method with PyTorch on NVIDIA GeForce RTX 4090 GPU.
In the training stage, we adopt Adam optimizer with an initial learning rate $1e-4$. 

\subsection{Comparision}

We compare our method with the representative denoising methods, including optimization-based denoisers
(Jet~\cite{cazals2005estimating}, GLR~\cite{zeng20193d}) and the recent deep-learning-based denoisers (PCN~\cite{rakotosaona2020pointcleannet}, DMR~\cite{Luo2020DifferentiableMR}, PF~\cite{zhang2020pointfilter}, PCDNF~\cite{liu2023pcdnf}, SBD~\cite{luo2021score}, PSR~\cite{chen2022deep})\color{rblue}{, PathNet~\cite{wei2024pathnet} and IterPFN~\cite{de2023iterativepfn}}. For those related works that share the same train-test split, i.e. SBD, PSR and IterPFN, \color{black}we use their pre-trained checkpoints of the network for the comparison experiments. \color{rblue}PCN, DMR, PF, PCDNF, and PathNet are re-trained with our training set using their released code. \color{black}We randomly sample 10K, 50K, and 100K points from each surface. Then, the clean point clouds are perturbed by isotropic Gaussian noise with standard deviation 1\%, 2\%, 3\% of the bounding sphere’s radius. These obtained noisy point clouds are saved as the input of all the methods for a fair comparison.

\noindent\textbf{Qualitative Performance.}
Our method outperforms the other related methods in terms of the visual quality of the denoised point clouds. Figure~\ref{fig:baseline} shows the denoising results of different methods on the point clouds with 3\% noise. The relatively large-scale noise makes it difficult to estimate the accurate underlying surface so many of these methods produce point clouds with remaining outliers and over-smoothed surfaces. 
Recent methods, such as PF, SBD, PSR, perform relatively clean results, but cannot deal with the shapes with thin structures such as the Casting model in the first row, and sharp features such as the Box Union model in the third row. By contrast, our method consistently produces high-quality denoised point clouds while maintaining the shape feature.

\begin{figure}[htbp]
    \centering
    \includegraphics[width=\linewidth]{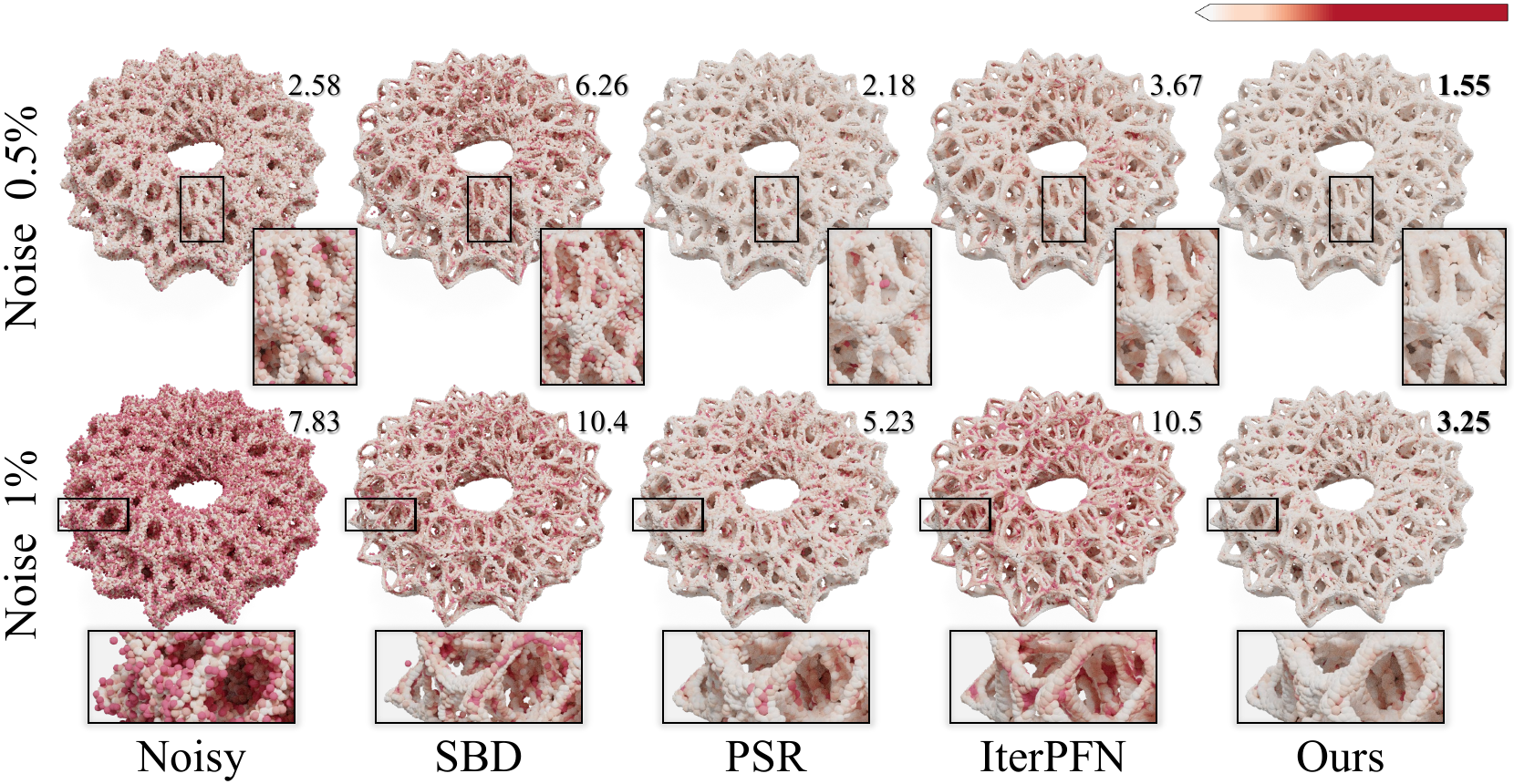}
    \vspace{-10pt}
    \caption{\textbf{Visual comparisons on complex Mobius model under $0.5\%$ and $1\%$ Gaussian noise.} We show the P2M ($\times10^5$) error of each result where red represents larger error values.}
    \label{fig:complex}
    \vspace{-20pt}
\end{figure}

We also present the results of these shapes with small-scale noise ($1\%$) in Figure~\ref{fig:detail}.
The optimization-based methods, i.e. Jet and GLR, produce highly over-smoothed results with missing shape details and sharp features, which can be clearly observed in the Box Union example in the fourth row. The results of PCN and DMR often have remaining noise and cause large shape errors compared to the ground-truth clean point cloud.
\color{rblue}{
IterPFN removes noise on the surface well but causes shrinkage when dealing with curved thin structures such as the Casting model in the first row.
}\color{black}
The other recent methods, including PF, PCDNF, SBD, PSR, and \color{rblue}{PathNet}\color{black}, obtain good denoised results with small shape errors, but have remaining noise on the plane surface such as the Box Union model in the third row, or over-smoothed surfaces at the sharp edges and detailed shapes such as the Quadric model in the last row and M1486 in the second row. By contrast, our method is comparably better at preserving shape details and sharp features.

\RE{In Figure~\ref{fig:complex}, we compare the iterative methods (SBD, PSR, IterPFN) in the denoised results of a complex shape with thin structures,}
coupled with $0.5\%$ and $1\%$. SBD \color{rblue}{and IterPFN} \color{black}produces denoised results with obvious shrinkage, which causes even larger shape errors than the input noisy point cloud. Comparing PSR and ours, we can produce smoother and cleaner results that preserve the complex shape details, while achieving smaller shape errors.

\begin{table*}[htbp]
    \centering
    \caption{\textbf{Quantitative evaluation of different methods on synthetic test sets.
    } CD and P2M are multiplied by $10^4$ and $10^5$, respectively.}
    \label{tab:baseline}
    \setlength{\tabcolsep}{5pt}
    \resizebox{\linewidth}{!}{
    \renewcommand\arraystretch{0.85}
    \begin{tabular}{c|lccccccccccc|ccccccccccc}
    \toprule
    \multicolumn{2}{c}{\# Metrics} & \multicolumn{11}{c}{CD ($10^4$)} & \multicolumn{11}{|c}{P2M ($10^5$)} \\
    \multicolumn{2}{c}{Noise} & \multicolumn{3}{c}{1\%}& & \multicolumn{3}{c}{2\%}& & \multicolumn{3}{c}{3\%} & \multicolumn{3}{|c}{1\%}& & \multicolumn{3}{c}{2\%}& & \multicolumn{3}{c}{3\%}\\
    \cmidrule{3-5}\cmidrule{7-9}\cmidrule{11-13}\cmidrule{14-16}\cmidrule{18-20}\cmidrule{22-24}
    Testset & Model                                      & 10K    & 50K    & 100K   & & 10K    & 50K    & 100K   & & 10K    & 50K    & 100K   & 10K    & 50K    & 100K   & & 10K    & 50K    & 100K   & & 10K    & 50K    & 100K      \\  
    \midrule 
    \multirow{11}{*}{\makecell{PU\\\cite{yu2018pu}}}              
                                & Jet       
                                                        & 3.19   & 0.98   & 0.70   & & 4.39   & 1.36   & 0.99   & & 5.43   & 2.17   & 1.58   & 10.4   & 3.39   & 3.24   & & 16.7   & 6.04   & 5.39   & & 24.1   & 12.3   & 10.2   \\    
                                & GLR       
                                                        & 2.97   & 0.67   & 0.42   & & 3.55   & 1.44   & 1.51   & & 4.71   & 2.43   & 2.79   & 10.7   & 1.66   & 1.33   & & 11.6   & 7.33   & 10.0   & & 20.8   & 15.4   & 21.1   \\    
                                \cmidrule{2-24}       
                                & PCN       
                                                        & 3.30   & 0.97   & 0.56   & & 6.26   & 1.24   & 0.75   & & 10.1   & 1.85   & 1.27   & 9.87   & 3.01   & 1.82   & & 29.3   & 4.53   & 3.08   & & 61.4   & 8.98   & 7.12   \\ 
                                & DMR       
                                                        & 4.44   & 1.18   & 0.88   & & 4.78   & 1.47   & 1.14   & & 5.51   & 2.17   & 1.79   & 16.8   & 4.79   & 4.65   & & 19.6   & 7.20   & 6.87   & & 25.6   & 13.1   & 12.5   \\ 
                                & PF       
                                                        & 3.41   & 0.74   & 0.46   & & 5.51   & 1.26   & 0.87   & & 7.39   & 1.68   & 1.24   & 10.9   & 1.56   & 1.21   & & 25.0   & 5.37   & 4.29   & & 40.8   & 8.15   & 7.06   \\ 
                                & PCDNF       
                                                        & 8.25   & 0.76   & 0.40   & & 9.11   & 1.22   & 0.66   & & 10.0   & 1.61   & \firc{0.92}   & 51.3   & 1.71   & 0.84   & & 54.6   & 4.97   & 2.60   & & 61.6   & 8.02   & \firc{4.58}   \\ 
                                & SBD       
                                                        & 2.52   & 0.71   & 0.42   & & 3.68   & 1.29   & 1.04   & & 4.71   & 1.93   & 1.75   & 4.63   & 1.50   & 1.12   & & 10.7   & 5.66   & 5.78   & & 19.4   & 10.4   & 10.8   \\   
                                & PSR       
                                                        & 2.42   & 0.65   & 0.37   & & 3.44   & 1.03   & 0.62   & & 4.17   & 1.40   & 1.06   & 3.31   & 0.75   & 0.49   & & 8.17   & 3.30   & 2.11   & & 13.4   & 5.82   & 4.93   \\ 
                                & IterPFN       
                                                        & \firc{2.05}   & 0.60   & 0.31   & & 3.04   & 0.80   & 0.55   & & 4.23   & 1.97   & 2.58   & 2.18   & 0.59   & 0.30   & & 5.55   & 1.82   & 1.80   & & 13.7   & 10.1   & 16.8   \\   
                                & PathNet       
                                                        & 2.79   & 0.68   & 0.38   & & 3.84   & 1.11   & 0.98   & & 4.64   & 2.37   & 3.26   & 7.30   & 1.05   & 0.67   & & 12.0   & 3.83   & 4.87   & & 17.6   & 13.2   & 22.9   \\ 
                                                        
                                \cmidrule{2-24}         
                                & \textbf{Ours}                          
                                                        & {2.09}   & \firc{0.59}   & \firc{0.30}   &\firc{} & \firc{2.99}   & \firc{0.78}   & \firc{0.50}   &\firc{} & \firc{3.62}   & \firc{1.18}   & {1.09}   & \firc{2.04}   & \firc{0.51}   & \firc{0.29}   &\firc{} & \firc{5.09}   & \firc{1.68}   & \firc{1.44}   &\firc{} & \firc{9.41}   & \firc{4.45}   & 5.51   \\  
    \hline\hline           
        \multirow{11}{*}{\makecell{PC\\\cite{rakotosaona2020pointcleannet}}}                    
                                & Jet       
                                                        & 3.53   & 1.21   & 0.83   & & 5.59   & 1.79   & 1.28   & & 7.21   & 2.63   & 1.84   & 10.1   & 2.58   & 1.98   & & 15.6   & 4.56   & 3.49   & & 22.2   & 6.85   & 4.60   \\    
                                & GLR       
                                                        & 4.36   & 1.01   & 0.51   & & 5.18   & 1.91   & 1.66   & & 7.46   & 3.23   & 3.08   & 13.9   & 1.89   & 0.73   & & 10.9   & 3.67   & 3.06   & & 19.2   & 7.05   & 8.36   \\    
                                \cmidrule{2-24}       
                                & PCN       
                                                        & 3.68   & 1.18   & 0.68   & & 7.99   & 1.63   & 0.92   & & 13.1   & 2.52   & 1.47   & 11.7   & 2.40   & 1.32   & & 28.0   & 3.86   & 1.95   & & 52.9   & 7.02   & 3.45   \\ 
                                & DMR       
                                                        & 6.52   & 1.59   & 1.05   & & 6.94   & 1.89   & 13.1   & & 7.71   & 2.68   & 2.03   & 21.5   & 3.89   & 2.46   & & 22.1   & 4.63   & 3.07   & & 24.5   & 7.26   & 5.72   \\ 
                                & PF       
                                                        & 4.43   & 1.03   & 0.67   & & 7.55   & 1.64   & 1.04   & & 9.84   & 2.41   & 1.44   & 19.2   & 2.06   & 1.31   & & 26.3   & 4.23   & 2.35   & & 39.3   & 7.54   & 3.48   \\ 
                                & PCDNF       
                                                        & 11.6   & 1.30   & 0.58   & & 13.2   & 1.89   & 0.90   & & 14.4   & 2.46   & 1.24   & 60.2   & 3.42   & 0.95   & & 64.3   & 5.69   & 1.93   & & 66.9   & 7.88   & 2.72   \\ 
                                & SBD       
                                                        & 3.37   & 1.07   & 0.57   & & 5.13   & 1.66   & 1.11   & & 6.78   & 2.49   & 1.91   & 8.30   & 1.77   & 0.81   & & 11.9   & 3.54   & 2.51   & & 19.4   & 6.57   & 5.13   \\   
                                & PSR       
                                                        & 3.14   & 0.99   & 0.51   & & 4.92   & 1.55   & 0.85   & & 6.14   & 2.12   & 1.27   & 10.1   & 1.67   & 0.65   & & 13.8   & 3.84   & 1.85   & & 17.8   & 5.92   & 2.88   \\   
                                & IterPFN       
                                                        & 2.62   & 0.91   & 0.46   & & 4.43   & 1.25   & 0.70   & & 6.02   & 2.54   & 2.23   & 6.98   & 1.39   & 0.61   & & 10.1   & 2.37   & 1.10   & & 15.6   & 7.15   & 5.87   \\   
                                & PathNet       
                                                        & 3.35   & 0.98   & 0.51   & & 5.13   & 1.46   & 0.99   & & 6.35   & 2.60   & 2.87   & 9.74   & 1.74   & 0.72   & & 13.6   & 3.06   & 1.86   & & 18.2   & 6.25   & 8.03   \\ 
                                                        
                                \cmidrule{2-24}         
                                & \textbf{Ours}                          
                                                        & \firc{2.59}   & \firc{0.89}   & \firc{0.43}   &\firc{} & \firc{4.41}   & \firc{1.24}   & \firc{0.67}   &\firc{} & \firc{5.56}   & \firc{1.93}   & \firc{1.23}   & \firc{6.33}   & \firc{1.06}   & \firc{0.33}   &\firc{} & \firc{9.43}   & \firc{2.11}   & \firc{0.82}   &\firc{} & \firc{13.2}   & \firc{4.21}   & \firc{2.13}   \\  
    \hline\hline           
        \multirow{11}{*}{\makecell{MOD\\\cite{huang2022modnet}}}                    
                                & Jet       
                                                        & 3.47   & 1.14   & 0.84   & & 5.36   & 1.80   & 1.42   & & 6.90   & 2.72   & 2.07   & 6.94   & 1.52   & 1.31   & & 9.81   & 2.68   & 1.75   & & 11.6   & 2.77   & 2.27   \\    
                                & GLR       
                                                        & 3.72   & 0.88   & 0.51   & & 4.64   & 1.69   & 1.61   & & 6.35   & 1.85   & 2.95   & 7.38   & 0.73   & 0.34   & & 6.09   & 1.18   & 1.14   & & 9.18   & 2.17   & 2.22   \\    
                                \cmidrule{2-24}       
                                & PCN       
                                                        & 3.34   & 1.05   & 0.63   & & 7.23   & 1.52   & 0.94   & & 12.1   & 2.42   & 1.60   & 3.64   & 0.91   & 0.48   & & 8.12   & 1.81   & 1.01   & & 13.0   & 2.92   & 1.59   \\ 
                                & DMR       
                                                        & 6.09   & 1.50   & 1.11   & & 6.55   & 1.90   & 1.46   & & 7.38   & 2.73   & 2.24   & 8.29   & 1.14   & 0.80   & & 8.53   & 1.60   & 1.15   & & 9.22   & 2.19   & 1.67   \\ 
                                & PF       
                                                        & 3.82   & 1.02   & 0.67   & & 6.50   & 1.65   & 1.07   & & 9.31   & 2.55   & 1.56   & 6.24   & 1.18   & 0.82   & & 12.3   & 2.69   & 1.85   & & 20.4   & 5.20   & 3.32   \\ 
                                & PCDNF       
                                                        & 9.31   & 1.10   & 0.55   & & 10.8   & 1.80   & 0.91   & & 12.2   & 2.54   & \firc{1.38}   & 11.9   & 1.10   & 0.37   & & 12.5   & 2.66   & 0.97   & & 13.9   & 4.38   & 1.75   \\ 
                                & SBD       
                                                        & 3.09   & 0.96   & 0.57   & & 4.75   & 1.64   & 1.22   & & 6.52   & 2.56   & 2.20   & 4.23   & 0.73   & 0.37   & & 5.46   & 1.40   & 0.99   & & 9.50   & 2.71   & 2.04   \\   
                                & PSR       
                                                        & 2.62   & 0.81   & 0.47   & & 4.28   & 1.33   & 0.86   & & 5.52   & 2.01   & {1.43}   & 3.90   & 0.51   & 0.28   & & 6.12   & 1.81   & 1.12   & & 8.16   & 3.21   & 1.71   \\   
                                & IterPFN       
                                                        & 2.41   & 0.77   & 0.44   & & 3.96   & 1.17   & 0.76   & & 5.65   & 2.50   & 2.72   & 3.05   & 0.46   & 0.25   & & 4.82   & 1.16   & 0.71   & & 5.83   & 1.84   & 1.95   \\   
                                & PathNet       
                                                        & 3.16   & 0.85   & 0.48   & & 4.81   & 1.41   & 1.10   & & 6.08   & 2.68   & 3.41   & 5.67   & 0.71   & 0.33   & & 8.09   & 1.44   & 0.90   & & 10.2   & 2.20   & 2.58   \\ 
                                                        
                                \cmidrule{2-24}         
                                & \textbf{Ours}                          
                                                        & \firc{2.35}   & \firc{0.75}   & \firc{0.42}   &\firc{} & \firc{3.90}   & \firc{1.15}   & \firc{0.73}   &\firc{} & \firc{5.06}   & \firc{1.77}   & {1.41}   & \firc{2.87}   & \firc{0.39}   & \firc{0.19}   &\firc{} & \firc{4.38}   & \firc{0.87}   & \firc{0.49}   &\firc{} & \firc{5.63}   & \firc{1.45}   & \firc{1.27}   \\  

    \bottomrule
    \end{tabular}
}
\end{table*}

\noindent\textbf{Quantitative Performance.} Table~\ref{tab:baseline} shows the quantitative evaluation of all the methods on the three synthetic test sets. Each method uses consistent hyper-parameters including trained network weights to produce the results of noisy point clouds with different densities and noise scales. Our method outperforms previous deep-learning-based methods and optimization-based methods quantitatively. The evaluation shows that our adaptive and iterative denoising approach consistently produces accurate clean point clouds with varying point densities and noise variation.

\subsection{Generalization}

\begin{figure*}[t]
    \centering
    \includegraphics[width=\linewidth]{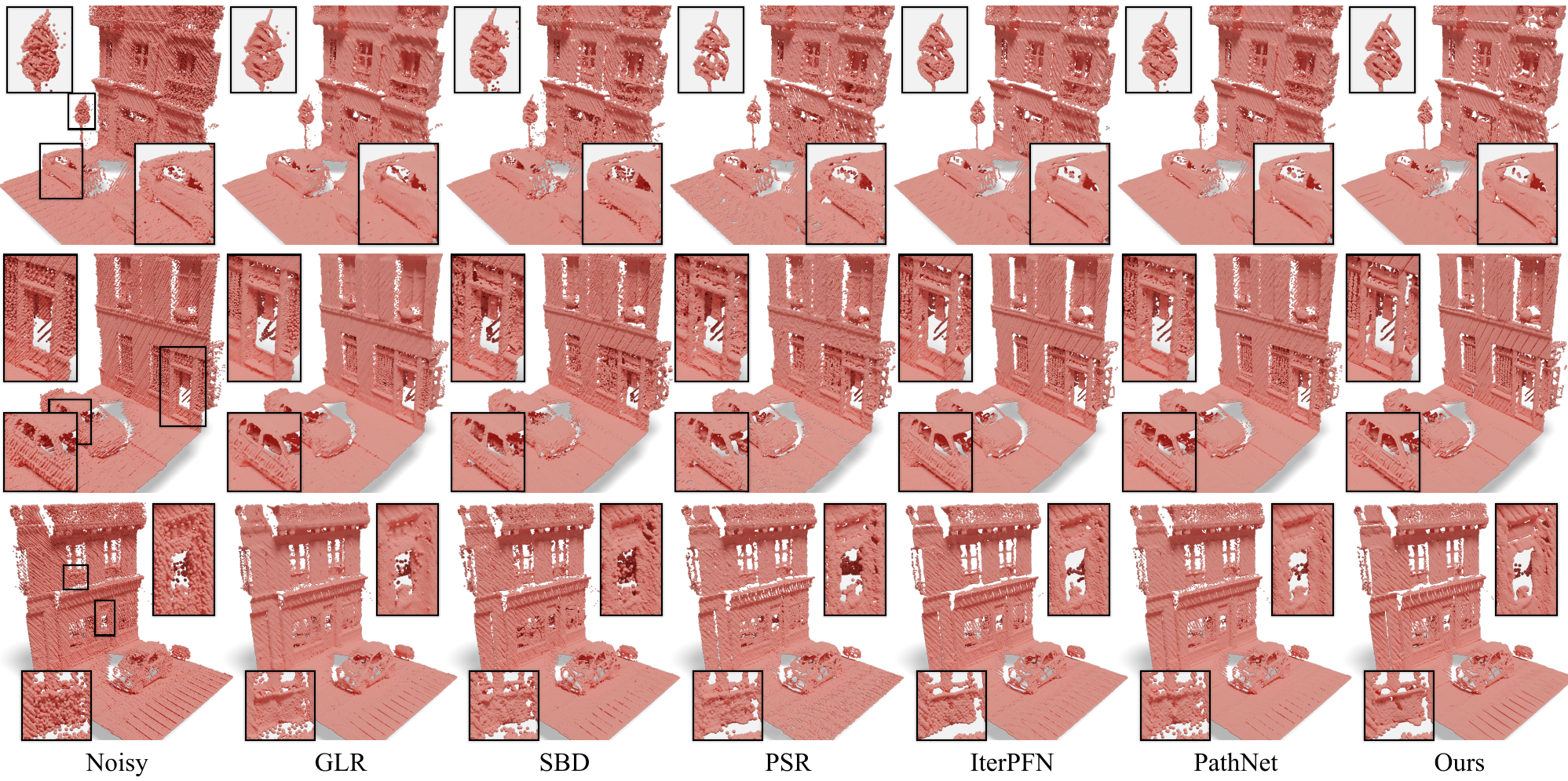}
    \vspace{-10pt}
    \caption{\textbf{Results on three scanned models from the Paris-rue-Madame Database \cite{serna2014paris}.} By contrast, our method obviously avoids shrinkage and produces smooth surfaces while preserving the shape details.}
    \label{fig:Ruemadame}
    \vspace{-10pt}
\end{figure*}
\begin{figure}[htbp]
    \centering
    \includegraphics[width=\linewidth]{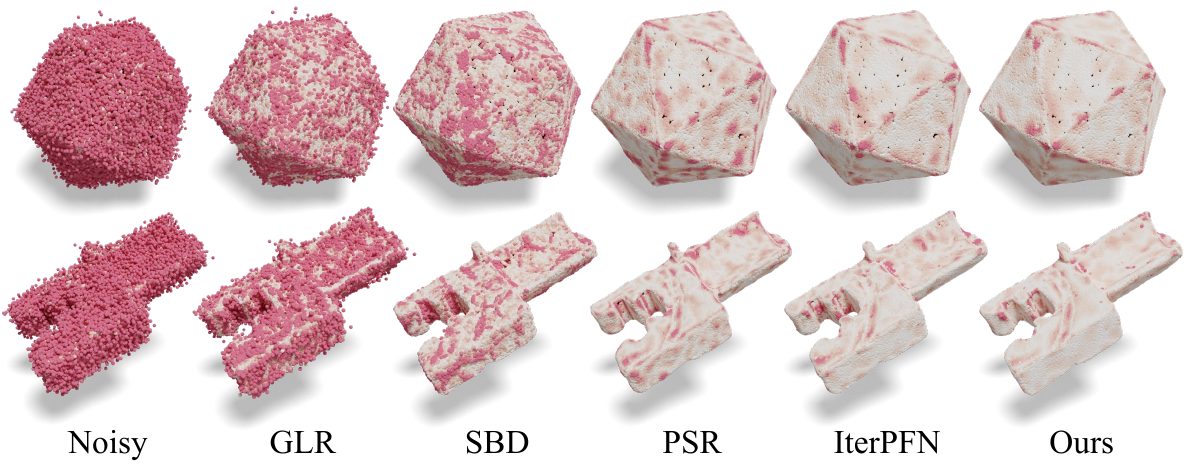}
    \vspace{-10pt}
    \caption{\textbf{Visual comparisons of denoising methods under $2\%$ Laplace noise.} Except for GLR, the other four methods are relatively robust to different noise patterns, but produce inaccurate results. By contrast, ours produces more accurate denoised results.}
    \label{fig:robustness_la}
    \vspace{-10pt}
\end{figure}

\noindent\textbf{More Noise Patterns.} 
To test the generalization ability of our method, we use the point clouds corrupted by other noise patterns as input and compare with some other methods (GLR, SBD, PSR\color{rblue}{, IterPFN}\color{black}) that perform well in the comparison experiments.
We consider the noise patterns including Laplace
noise (LA), discrete noise (DC), anisotropic Gaussian noise (AG), uni-directional Gaussian noise (UG) and uniform noise (UF). Note that none of the above methods have been trained on these noise patterns. 

Figure~\ref{fig:robustness_la} and Table~\ref{tab:robust} report the qualitative and quantitative evaluations. Although we have tweaked the hyper-parameters of GLR carefully, they still result in dramatically decreasing performance on the Laplacian noise. Similarly, the results of SBD also have remaining outliers around the denoised point cloud. On the other hand, although the PSR\color{rblue}{, IterPFN} \color{black} and ours produce visually similar results, our method exhibits much better performance in quantitative evaluation. It demonstrates the superiority of our adaptive and iterative denoising strategy in dealing with various noise patterns.

\begin{table}[htbp]
    \centering
    \caption{\textbf{Comparison among competitive denoising methods under different patterns of noise.} P2M is multiplied by $10^5$.}
    \label{tab:robust}
    \setlength{\tabcolsep}{3pt}
    \resizebox{\linewidth}{!}{
    \renewcommand\arraystretch{1}
    \begin{tabular}{c|lccccccccccc}
    \toprule
    \multicolumn{2}{c}{PU-Set} & \multicolumn{3}{c}{1\%}& & \multicolumn{3}{c}{2\%}& & \multicolumn{3}{c}{3\%} \\
    \cmidrule{3-5}\cmidrule{7-9}\cmidrule{11-13}
     Type & Model                                        & 10K    & 50K    & 100K   & & 10K    & 50K    & 100K   & & 10K    & 50K    & 100K    \\  
    \midrule 
    \multirow{6}{*}{\makecell{LA}}              

                                & GLR       
                                                         & 20.5   & 3.67   & 2.81   & & 20.4   & 21.6   & 28.1   & & 41.2   & 44.8   & 59.9       \\   
                                & SBD       
                                                         & 6.66   & 2.27   & 1.68   & & 18.2   & 8.71   & 9.46   & & 33.9   & 16.6   & 19.3       \\    
                                & PSR       
                                                         & 4.97   & 1.34   & 0.85   & & 12.1   & 5.01   & 3.14   & & 19.8   & 8.86   & \firc{8.93}       \\
                                & IterPFN       
                                                         & 3.15   & 0.85   & 0.48   & & 8.05   & 3.09   & 4.42   & & 31.8   & 24.1   & 35.3       \\

                                \cmidrule{2-13}
   
                                & \textbf{Ours}
                                                         & \firc{3.08}   & \firc{0.83}   & \firc{0.47}   &\firc{} & \firc{7.97}   & \firc{2.90}   & \firc{2.25}   &\firc{} & \firc{15.4}   & \firc{8.47}   & {10.4}       \\
    \hline
    \multirow{6}{*}{\makecell{DC}}              

                                & GLR       
                                                         & 18.5   & 1.43   & 0.45   & & 11.1   & 1.48   & 0.99   & & 20.4   & 3.30   & 2.60       \\   
                                & SBD       
                                                         & 2.46   & 0.48   & 0.35   & & 4.19   & 1.37   & 1.03   & & 9.76   & 2.45   & 1.97       \\    
                                & PSR       
                                                         & 1.89   & 0.28   & 0.18   & & 2.96   & 0.62   & 0.47   & & 4.71   & 2.06   & 3.02       \\
                                & IterPFN       
                                                         & 0.87   & 0.15   & 0.09   & & 1.83   & \firc{0.37}   & \firc{0.22}   & & 3.09   & \firc{0.77}   & \firc{0.57}       \\

                                \cmidrule{2-13}
   
                                & \textbf{Ours}
                                                         & \firc{0.84}   & \firc{0.13}   & \firc{0.08}   &\firc{} & \firc{1.78}   & {0.43}   & {0.31 }   & & \firc{2.89}   & {0.95}   & 0.85       \\
    \hline
    \multirow{6}{*}{\makecell{AG}}              

                                & GLR       
                                                         & 19.6   & 2.50   & 1.34   & & 13.5   & 7.51   & 10.2   & & 27.0   & 16.1   & 21.7       \\   
                                & SBD       
                                                         & 4.55   & 1.49   & 1.12   & & 11.0   & 5.94   & 6.18   & & 19.9   & 11.7   & 12.2       \\    
                                & PSR       
                                                         & 3.41   & 0.76   & 0.52   & & 8.42   & 3.44   & 2.27   & & 14.0   & 6.86   & \firc{6.58}       \\
                                & IterPFN       
                                                         & 2.06   & 0.61   & 0.32   & & 5.63   & 2.02   & 2.13   & & 17.1   & 14.0   & 19.1       \\

                                \cmidrule{2-13}
   
                                & \textbf{Ours}
                                                         & \firc{2.04}   & \firc{0.53}   & \firc{0.30}   &\firc{} & \firc{5.45}   & \firc{1.97}   & \firc{1.78}   &\firc{} & \firc{10.3}   & \firc{6.04}   & {7.95}       \\
    \hline
    \multirow{6}{*}{\makecell{UG}}              

                                & GLR       
                                                         & 18.9   & 1.73   & 0.71   & & 12.1   & 2.45   & 2.42   & & 22.6   & 5.97   & 5.81       \\   
                                & SBD       
                                                         & 2.78   & 0.65   & 0.48   & & 5.45   & 2.43   & 2.10   & & 11.4   & 4.80   & 4.30       \\    
                                & PSR       
                                                         & 2.29   & 1.12   & 1.75   & & 4.04   & 1.91   & 1.86   & & 6.92   & 3.59   & {3.28}       \\    
                                & IterPFN       
                                                         & \firc{1.03}   & 0.21   & 0.14   & & \firc{2.40}   & {1.01}   & 0.69   & & 7.20   & 4.86   & 4.51       \\

                                \cmidrule{2-13}
   
                                & \textbf{Ours}
                                                         & 1.07   & \firc{0.20}   & \firc{0.12}   & & {2.56}   & \firc{0.98}   & \firc{0.65}   &\firc{} & \firc{5.26}   & \firc{2.93}   & \firc{2.87}       \\
    \hline
    \multirow{6}{*}{\makecell{UF}}              

                                & GLR       
                                                         & 18.4   & 1.42   & 0.45   & & 11.1   & 1.28   & 0.89   & & 20.4   & 3.52   & 2.46       \\   
                                & SBD       
                                                         & 2.49   & 0.47   & 0.31   & & 4.10   & 1.28   & 1.03   & & 10.2   & 2.96   & 2.08       \\    
                                & PSR       
                                                         & 1.92   & 0.27   & 0.17   & & 3.01   & 0.63   & 0.54   & & 4.77   & 2.36   & 3.32       \\    
                                & IterPFN       
                                                         & 0.87   & 0.15   & 0.09   & & 1.93   & 0.53   & 0.28   & & 3.41   & 1.10   & 0.71       \\

                                \cmidrule{2-13}
   
                                & \textbf{Ours}
                                                         & \firc{0.84}   & \firc{0.13}   & \firc{0.08}   &\firc{} & \firc{1.87}   & \firc{0.43}   & \firc{0.24}   &\firc{} & \firc{3.08}   & \firc{0.87}   & \firc{0.65}       \\

    \bottomrule
    \end{tabular}
}
\end{table}

\noindent\textbf{Real-Scan Denoising.} 
Figure~\ref{fig:Ruemadame} shows the denoised results of different methods on the real-scanned Paris-rue-Madame dataset~\cite{serna2014paris}. We zoom in some of the regions to emphasize the different performances of these methods. As shown with the tree example in the first row, GLR and SBD often produce remaining outliers after denoising. PSR produces much cleaner results but sometimes causes larger holes in the results, such as in the tree example. 
\color{rblue}{
IterPFN and PathNet protect the overall shape better, but cannot recover the geometric details clearly, such as the sill in the last row.
}\color{black}
By contrast, our method can not only generate clearer and smoother point clouds on the surface but also preserve subtle structures and geometric details. As a result, our denoised points are scattered on the underlying surface uniformly and smoothly.

\subsection{Ablation Study}

We conduct the ablation study from two aspects. One is to validate the effectiveness of using the adaptive and iterative denoising strategy. We compare with the variants of our method without the iterative process or without the adaptive schedule. The other is to validate our key designs to strengthen the performance of our adaptive and iterative denoising process. Our key designs include adopting the diffusion model for point cloud denoising, the feature fusion, and the gradient fusion module in the denoising network.

\subsubsection{Effectiveness of Adaptive and Iterative Schedule}

To validate the effectiveness of our adaptive and iterative denoising strategy, we construct two variants of our method to make the comparison, as shown in the first two rows in Table~\ref{tab:ablation_study}. The first variant, denoted as OneStep, is a non-adaptive and non-iterative variant, which directly uses the network to predict the score vectors from the input noisy point cloud, and updates the point clouds only once to achieve the denoised result. Because it is difficult to predict the accurate score vectors from the noisy point clouds, the quantitative results are obviously worse than other iterative baselines.

The second variant, denoted as FixedSched, is iterative but non-adaptive.
Specifically, with our trained network, we pre-define a fixed iterative schedule following $\{\alpha_t=\alpha(\alpha_{decay})^t\}^T_{t=1}$, as did in SBD~\cite{luo2021score}.
Based on our experiment results, it is able to improve the denoised results, but it is difficult to find a common fixed schedule for all the settings, i.e. point cloud density and noise scale. A long iterative process with many steps can alleviate this problem, but it requires much more running time cost. For example, we achieve comparable results with a 30-step fixed schedule to our results with a 5-step adaptive schedule. We report the quantitative results with 5-step and 30-step fixed schedules in Table~\ref{tab:ablation_study}, denoted as FixedSched5 and FixedSched30 respectively.

\begin{table*}[t]
    \centering
    \caption{\textbf{Ablation study results.} The advantage of adaptive denoising schedule is demonstrated by comparing it to the OneStep and FixedSched baselines. The effectiveness of our key designs is validated by respectively adopting the generative diffusion model (GDM), removing feature fusion (w/ $F_T$, w/ $F_t$, w/ $F_{t+T}$), removing gradient fusion (w/o GF and w/ GF($const$)). CD and P2M are multiplied by $10^4$ and $10^5$.}
    \label{tab:ablation_study}
    \resizebox{\linewidth}{!}{
    \renewcommand\arraystretch{1}
    \begin{tabular}{lccccccccccc|ccccccccccc}
    \toprule
           & \multicolumn{11}{c}{CD ($10^4$)} & \multicolumn{11}{|c}{P2M ($10^5$)} \\
    PU-Set & \multicolumn{3}{c}{1\%}& & \multicolumn{3}{c}{2\%}& & \multicolumn{3}{c}{3\%} & \multicolumn{3}{|c}{1\%}& & \multicolumn{3}{c}{2\%}& & \multicolumn{3}{c}{3\%}\\
    \cmidrule{2-4}\cmidrule{6-8}\cmidrule{10-12}\cmidrule{13-15}\cmidrule{17-19}\cmidrule{21-23}
     Ablation & 10K    & 50K    & 100K   & & 10K    & 50K    & 100K   & & 10K    & 50K    & 100K   & 10K    & 50K    & 100K   & & 10K    & 50K    & 100K   & & 10K    & 50K    & 100K\\  
    \midrule          
        OneStep
        & 2.17   & 0.61   & 0.33   & & 3.15   & 0.88   & 0.62   & & 3.86   & 1.53   & 1.54   & 2.21   & 0.59   & 0.38   & & 6.52   & 2.32   & 2.36   & & 12.4   & 7.16   & 9.67   \\ 
        FixedSched5
        & 2.18   & 0.62   & 0.34   & & 3.12   & 0.85   & 0.57   & & 3.75   & 1.27   & 1.34   & 2.24   & 0.65   & 0.33   & & 6.08   & 2.15   & 1.84   & & 10.5   & 5.01   & 7.72   \\ 
        FixedSched30
        & 2.16   & 0.61   & \secc{0.32}   & & 3.09   & 0.83   & \secc{0.51}   & & 3.70   & 1.23   & \firc{1.05}   & 2.28   & 0.58   & 0.34   & & 5.62   & 1.89   & 1.53   & & 9.97   & 4.76   & \firc{5.17}   \\ 
    \midrule
        GDM
        & 2.75   & 0.64   & 0.37   & & 3.32   & 1.01   & 0.76   & & 4.08   & 1.74   & 1.67   & 2.71   & 0.69   & 0.64   & & 6.41   & 3.42   & 3.65   & & 12.4   & 9.09   & 10.8   \\ 
        w/ $F_{T}$
        & 2.15   & 0.61   & 0.33   & & \secc{3.04}   & \secc{0.81}   & {0.52}   & & \secc{3.65}   & \secc{1.20}   & 1.10   & 2.19   & 0.58   & 0.34   & & 5.43   & {1.79}   & \secc{1.48}   & & {9.70}   & \secc{4.52}   & 5.37   \\ 
        
        w/ $F_t$
        & {2.13}   & \secc{0.60}   & {0.32}   & & 3.06   & 0.83   & 0.57   & & 3.70   & 1.47   & 1.48   & {2.12}   & \secc{0.52}   & \secc{0.31}   & & {5.32}   & 2.04   & 1.93   & & 9.98   & 6.65   & 8.87   \\ 
        
        w/ $F_{t+T}$
        & \secc{2.11}   & \secc{0.60}   & \secc{0.31}   & & 3.05   & \secc{0.81}   & \secc{0.51}   & & 3.68   & 1.26   & 1.22   & \secc{2.10}   & 0.54   & 0.35   & & \secc{5.29}   & \secc{1.76}   & 1.51   & & \secc{9.54}   & 4.82   & 6.83   \\ 
        
        w/o GF
        & 2.22   & 0.61   & 0.33   & & 3.16   & 0.87   & 0.54   & & 3.86   & 1.34   & 1.24   & 2.83   & 0.68   & 0.38   & & 6.56   & 2.19   & 1.63   & & 11.3   & 5.37   & 6.61   \\ 
        w/ GF($const$)
        & 2.16   & \secc{0.60}   & {0.32}   & & \secc{3.04}   & 0.83   & {0.52}   & & 3.67   & 1.25   & 1.13   & 2.22   & 0.58   & 0.35   & & 5.48   & 1.98   & 1.54   & & 9.76   & 4.96   & 5.76   \\ 

    \midrule
        Full
        & \firc{2.09}   & \firc{0.59}   & \firc{0.30}   &\firc{} & \firc{2.99}   & \firc{0.78}   & \firc{0.50}   &\firc{} & \firc{3.62}   & \firc{1.18}   & \secc{1.09}   & \firc{2.04}   & \firc{0.51}   & \firc{0.29}   & & \firc{5.09}   & \firc{1.68}   & \firc{1.44}   &\firc{} & \firc{9.41}   & \firc{4.45}   & \secc{5.51}   \\  
    \bottomrule
    \end{tabular}
}
\end{table*}

In contrast, the full implementation of our method uses the adaptive schedule of our formulation of the score-based diffusion model. It produces obviously better-denoised point clouds in Table~\ref{tab:ablation_study}, especially for the large-scale noises, validating the advantage of the adaptive schedule in the iterative denoising process.

\subsubsection{Key Designs in Score-based Point Cloud Denoising}

We conduct the following experiments to validate the key designs, i.e. point cloud denoising diffusion, feature fusion, and gradient fusion, in each iteration step to preserve the detailed shape features of the underlying surface during denoising. The quantitative evaluations of these experiments are listed in Table~\ref{tab:ablation_study}.

\color{rred}\noindent\textbf{Diffusion Model for Point Cloud Denoising.} As explained in Section~\ref{sec:diffusion}, we use the tailored score-based diffusion model for the point cloud denoising task (Equation~\ref{eq:diffusion_equation}) rather than the conventional generative diffusion model. To validate this, we conduct the alternative experiment, denoted as GDM (i.e. Generative Diffusion Model) in Table~\ref{tab:ablation_study}\color{black}, where we construct the noisy point cloud during training based on the original DDPM formulation instead of our Equation~\ref{eq:diffusion_equation}. As for the iterative denoising process, the sampling formulation, i.e. Equation~\ref{eq:sampling}, is replaced by
\begin{equation}
    x^{t-1}=\frac{1}{\sqrt{\alpha_t}}\left(x^t-\frac{\beta_t}{\sqrt{1-\bar\alpha_t}}s_\theta(x^t)\right)+z,\ z\sim\mathcal N(0,\ \beta_t{\rm I})
\end{equation}
Note that it prevents the use of our feature fusion due to the reason that we can only fuse the point-wise features of the point clouds at the same scale. So we directly use the shape features of the original noisy point cloud to predict the score vectors without the feature fusion technique. Consequently, the denoising performance decreases dramatically, validating the importance of the modified diffusion model formulation for point cloud denoising.

\begin{figure}[tbp]
    \centering
    \includegraphics[width=\linewidth]{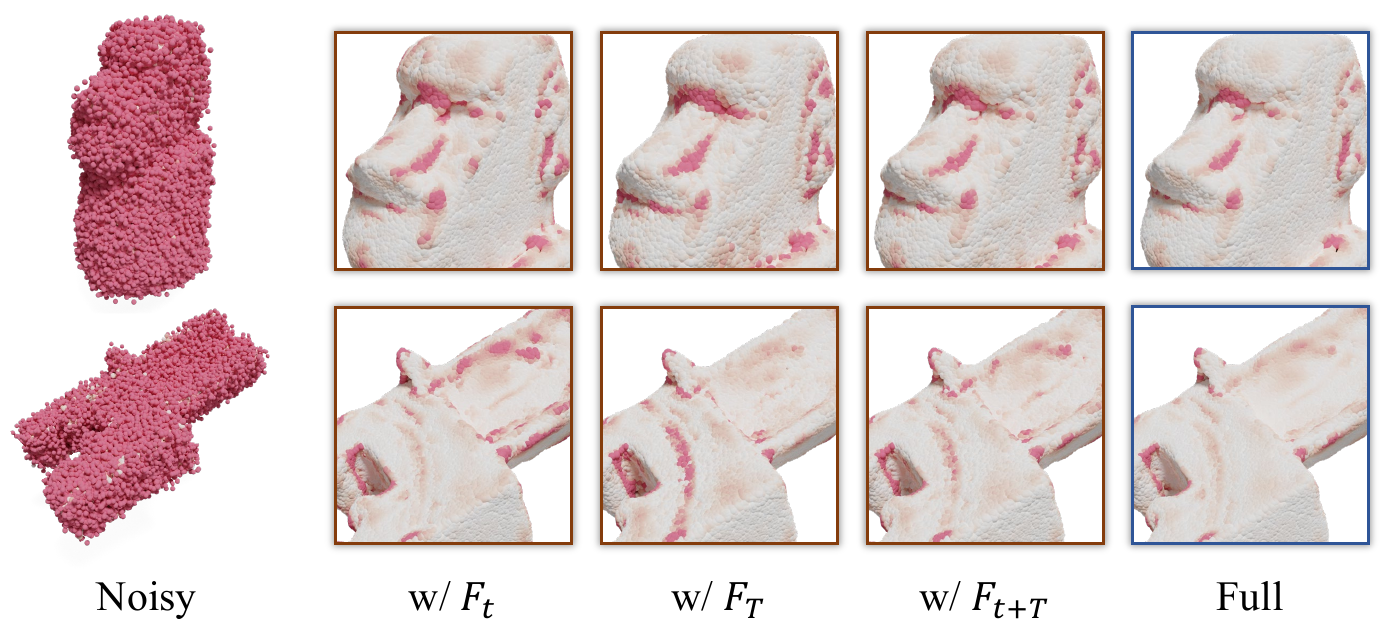}
    \vspace{-15pt}
    \caption{\textbf{Shape-preserving performance with or without the feature fusion module in our network.} 
    }
    \vspace{-20pt}
    \label{fig:more_noise_model}
\end{figure}

\noindent\textbf{Feature Fusion.}
\color{rbrown}Without the feature fusion module in our network, there are three alternative ways to select the point-wise features as input of the gradient prediction module. \color{black}The first is to use the shape features of the original noisy point cloud $x$ at the very beginning, as did in the SBD approach~\cite{luo2021score}, denoted as w/ $F_T$. The second is to use the shape feature of the updated point cloud $x^{t}$ at each step for score prediction, denoted as w/ $F_t$. \color{rbrown}The last one is to use the mean of the features of the current and original noisy point clouds, denoted as w/ $F_{t+T}$.  \color{rbrown}We retrain the network for each of the experiments. \color{black}
Table~\ref{tab:ablation_study} demonstrates that the shape feature of the updated point clouds often obtains slightly worse quantitative results under the large noise, since it may lose the shape features during the process. 
As shown in Figure~\ref{fig:more_noise_model}, using the features of updated point clouds, i.e. w/ $F_t$, produces relatively smooth denoised point clouds under most surfaces, 
\color{rbrown}{but leaves obvious detail errors, since there is a lack of guidance from the shape feature of original point cloud. 
}\color{black}
By contrast, the feature fusion in our method aggregates the information during the iterative process and generates better-denoised point clouds while preserving the shape features.

\begin{figure}[t]
    \centering
    \includegraphics[width=\linewidth]{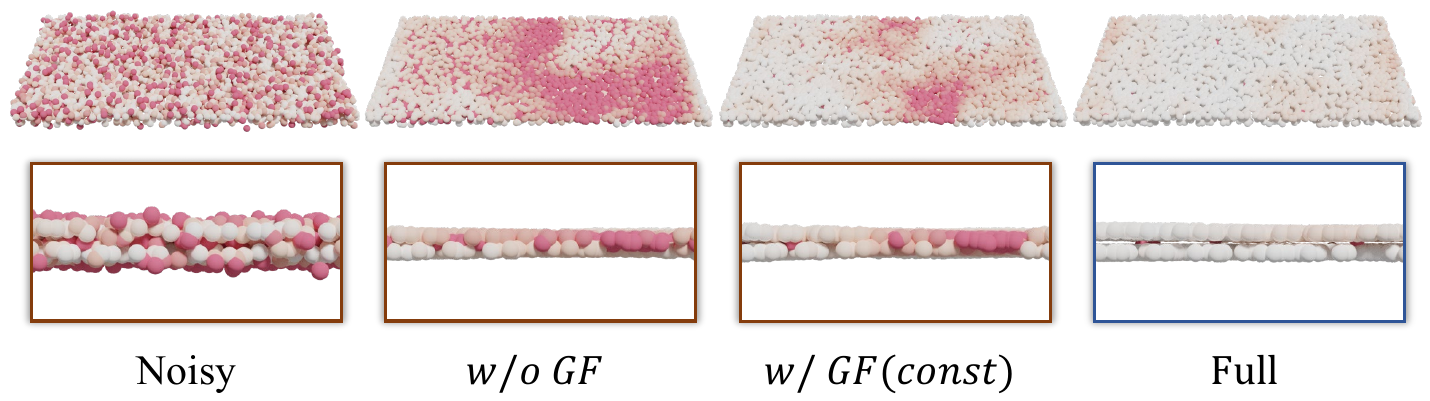}
    \vspace{-5pt}
    \caption{\textbf{Thin-structure preserving performance with and without gradient fusion module in our network.} We sample 10k points from a thin plane mesh and add 1\% Gaussian noise. By contrast, our full method can protect the thin structure better, while $w/o GF$ and $w/ GF(const)$ lead to surface degradation.}
    \label{fig:thin_struture}
    \vspace{-20pt}
\end{figure}

\noindent\textbf{Gradient Fusion.}
We use two different implementations to remove our gradient fusion module. One is denoted as w/o GF with $k=1$, i.e. querying only one nearest neighbor from $x^t$ and take its point-wise feature for score prediction. \color{rbrown}The other uses the average gradient fusion, denoted as w/ GF($const$). \color{black}Again, as shown in Table~\ref{tab:ablation_study}, they cause a performance decrease due to the inaccurate score prediction, especially for the sparse point clouds.
\color{rbrown}{
Figure~\ref{fig:thin_struture} presents the visual results of these three strategies. It demonstrates that our strategy is superior in protecting the thin structures, since the local importance weight assists in separating close surfaces.
}\color{black}

\subsection{Hyperparameters}\label{sec:hyperparmeter}

In the above experiments, we empirically select the proper hyper-parameters. Here we report the experiment results to validate the selection of the hyper-parameters. It demonstrates that the selected hyper-parameters are robust to different settings of the noisy point clouds, including the point densities and noise scales.


\noindent\textbf{Iteration Number $L$.} The formulation of the score-based diffusion model allows us to compute an adaptive schedule for the iterative denoising of varying noisy point clouds. Once the iteration number $L$ is specified, we only need to estimate the total noise variance and determine the schedule $\{\tau_l\}$ based on Equation~\ref{eq:optimalbeta}.

\begin{figure}[t]
        \centering
        \includegraphics[width=\linewidth]{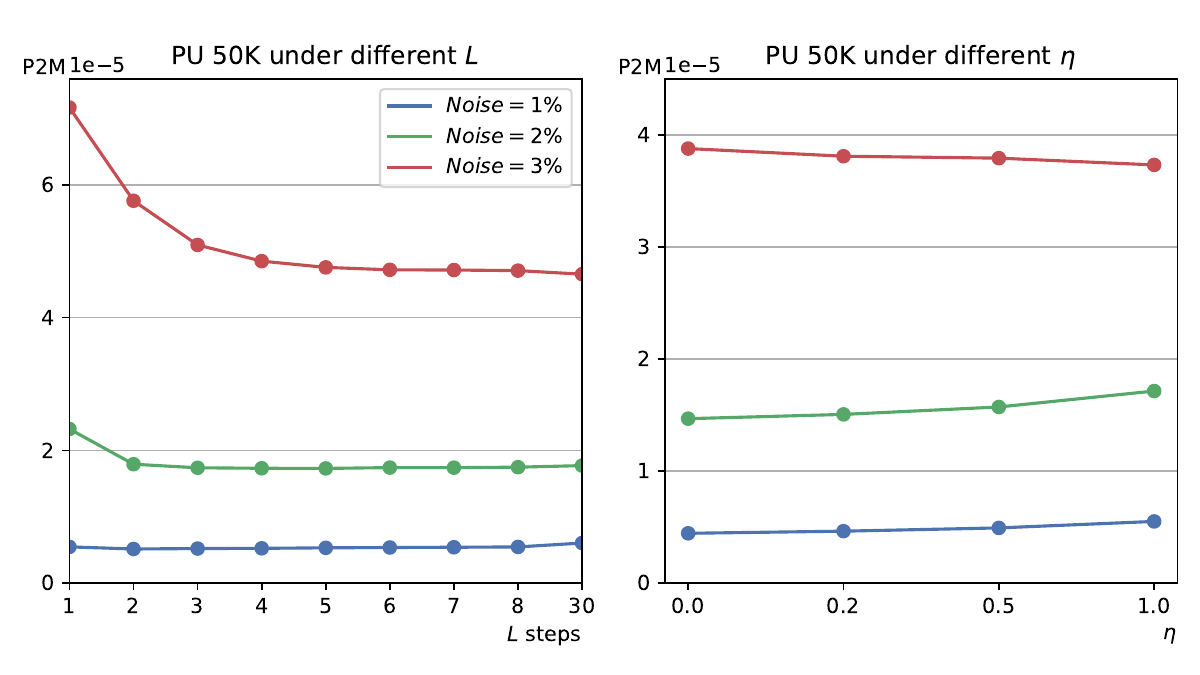}
        \vspace{-15pt}
        \caption{\textbf{The denoising performance with adaptive schedules arranged with different numbers of timesteps $L$ and $\eta$.} The performance improvement is negligible when $L$ is larger than 5. }
        \label{fig:TandNiter}
        \vspace{-10pt}
\end{figure}

The left picture of Figure~\ref{fig:TandNiter} presents the denoising performance measured with P2M metric when setting different numbers of iterations during denoising. We tested $L$ ranging from 1 to 8, as well as the case $L=30$. It shows that the denoising performance becomes stable as long as the iteration number is larger than 5. More iteration steps don't improve the performance obviously and would cause more running time costs. Therefore, we set $L=5$ in all the other experiments for efficiency purposes.

\begin{figure}[t]
    \centering
    \includegraphics[width=\linewidth]{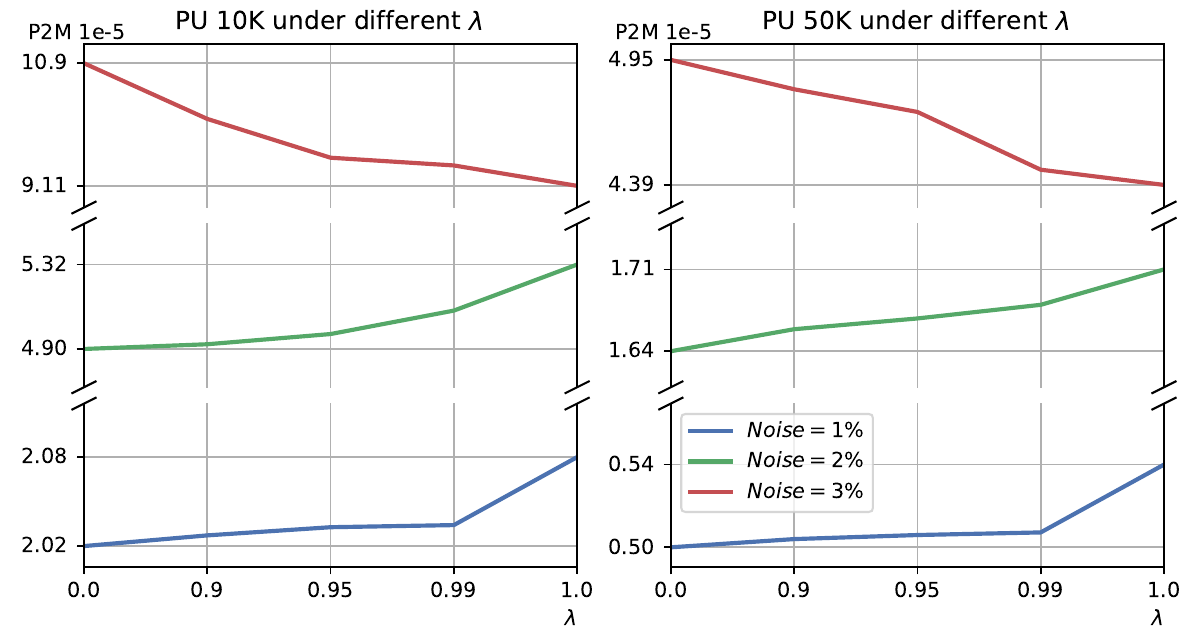}
    \vspace{-15pt}
    \caption{\textbf{The denoising performance with varying $\lambda$.} We select $\lambda$ for the noisy point clouds with different scales of noise.}
    \label{fig:lambda}
    \vspace{-20pt}
\end{figure}

\noindent\textbf{Sampling Noise Weight $\eta$.}
The score-based diffusion model for point cloud denoising uses estimated noise to sample the denoised points in the next step. The sampling is illustrated in Equation~\ref{eq:mu_theta} and~\ref{eq:sampling_org}, where the variable $\eta$ controls whether the noise variance $\Sigma$ is used in each iteration step. In other words, $\eta=0$ uses a deterministic sampling process to iteratively update the point clouds.

We test the performance with $\eta \in [0,1]$ in the right picture of \ref{fig:TandNiter}. For small-scale noise, such as $1\%$ and $2\%$ noises, the results are slightly better when $\eta=0$, while for large-scale $3\%$ noise the performance is superior when $\eta=1$. We assume the reason as the predicted scores of the trained network are not accurate when the point cloud is coupled with large-scale noise, and inject a small amount of random noise helps to generate reasonable results. But overall, the quantitative results with varying $\eta$ are close to each other. We set $\eta=0$ in all the other experiments for improving the accuracy at small-level noise.



\noindent\textbf{Loss Weight $\lambda$.} The variable $\lambda$ controls whether our network focuses more on the small-scale noise or the large-scale noise during training. The network has a larger weight on the loss of those timesteps with small-scale noise when $\lambda=0$, and the opposite when $\lambda=1$. Figure~\ref{fig:lambda} shows how the denoising performance varies along with $\lambda$ on different scales of noises. We set $\lambda=0.99$ in our experiments for a balanced performance.

\section{Conclusion}
We propose an adaptive and iterative point cloud denoising method with the score-based diffusion model. For each given noisy point cloud, we estimate its noise variance and determine an adaptive schedule for the iterative denoising purpose. We design and train the denoising network architecture with feature fusion and gradient fusion to recover the underlying surface. Our method effectively reduces noise and produces clean point clouds, alleviating the shrinkage and outlier problems.

However, the denoised point clouds of our method are still not exactly distributed on the underlying surface, especially for the sparse point cloud with large-scale noise. It is difficult to estimate the accurate surface based on merely the noisy point cloud, when the network is trained on a dataset with a limited amount of 3D data. Exploiting the 3D prior with a large-scale dataset or lifting the 2D prior with state-of-the-art foundation models would further promote the point cloud denoising performance.

\noindent\textbf{Acknowledgments}

\noindent
We thank the anonymous reviewers for their helpful comments and suggestions. This work is supported by the Joint Funds of the National Natural Science Foundation of China (U23A20312), the National Key R\&D Program of China (2021YFB1715900), the National Natural Science Foundation of China (62272277, 62072284, 62302269), and a grant from the Natural Science Foundation of Shandong Province (No. ZR2023QF077). Manyi Li is supported by the Excellent Young Scientists Fund Program (Overseas) of Shandong Province (Grant No.2023HWYQ-034).

\noindent\textbf{Data Citation}

\noindent
The data that support the findings of this study are available from the corresponding author upon reasonable request.

\bibliographystyle{eg-alpha-doi} 

\bibliography{refer}

\newpage
\appendix

\section{Derivation in the forward diffusion process}

As described in Section~\ref{sec:diffusion}, noisy point clouds' conditional distribution can be considered a Markov chain. Here we explain the derivation of Equation~\ref{eq:diffusion_equation} in Section~\ref{sec:diffusion}. Considering that the Gaussian distribution is additive, we can expand $q(x^t|x^{0})$ as follows:
\begin{equation}
\begin{aligned}
x^{t}&=x^0+\sqrt{\frac{1-\bar\alpha_t}{\bar\alpha_t}}z,&z\sim \mathcal{N}({\rm 0},{\rm I}),\\
&=x^{t-1}-\sqrt{\frac{1-\bar\alpha_{t-1}}{\bar\alpha_{t-1}}}z_{t-1}+\sqrt{\frac{1-\bar\alpha_t}{\bar\alpha_t}}z_t,&z_t,z_{t-1}\sim \mathcal{N}({\rm 0},{\rm I}),\\
&=x^{t-1}+\sqrt{\frac{{\beta_t}}{{\bar\alpha_t}}}z,&z\sim \mathcal{N}({\rm 0},{\rm I}).
\end{aligned}
\end{equation}
Its distribution form is described as follows:
\begin{equation}\label{eq:x0toxt}
    q(x^t|x^{t-1})=\mathcal{N}(x^t;x^{t-1},\beta_t/\bar{\alpha}_t{\rm I}).
\end{equation}

\section{Derivation in the sampling process}

According to DDPM~\cite{ho2020denoising}, the conditional distribution $p_\theta(x^{t-\Delta}|x^t)=\mathcal{N}(x^{t-\Delta};\mu_\theta,\Sigma)$ of sampling process aims to restore the sequence from $x^{t}$ to $x^{t-\Delta}$, \color{rbrown}where $\Delta\in[1,t]$\color{black}. For the corresponding diffusion process, this means the $p_\theta(x^{t-\Delta}|x^t)$ and $q(x^{t-\Delta}|x^t)=\mathcal{N}(x^{t-\Delta};\tilde\mu_t,\tilde{\sigma}^2_t\ {\rm I})$ distributions are equivalent in definition, where $q(x^{t-\Delta}|x^t)$ can be formed as $q(x^{t-\Delta}|x^t, x^0)$. 

\noindent Based on Bayes' theorem, we can expand the distribution as follows:
\begin{equation}\color{rbrown}{
\begin{aligned}
q(x^{t-\Delta}|x^{t},x^0)&={\mathcal N(x^{t};x^{t-\Delta},\frac{\bar\alpha_{t-\Delta}-\bar\alpha_{t}}{\bar\alpha_{t-\Delta}\bar\alpha_{t}}{\rm I})}\frac{\mathcal N(x^{t-\Delta};x^{0},\frac{1-\bar\alpha_{t-\Delta}}{\bar\alpha_{t-\Delta}}{\rm I})}{\mathcal N(x^{t};x^{0},\frac{1-\bar\alpha_{t}}{\bar\alpha_t}{\rm I})}\\
&\propto \exp(-\frac 1 2({\frac{\bar\alpha_{t}\bar\alpha_{t-\Delta}(x^t-x^{t-\Delta})^2}{\bar\alpha_{t-\Delta}-\bar\alpha_{t}}}+{\frac{\bar\alpha_{t-\Delta}(x^{t-\Delta}-x^{0})^2}{1-\bar\alpha_{t-\Delta}}}\\
&\quad\quad\quad\quad\quad\quad\quad\quad\quad\quad\ \ \ -{\frac{\bar\alpha_{t}(x^t-x^{0})^2}{1-\bar\alpha_t}}))\\
&= \exp(-\frac 1 2(\underbrace{\frac {\bar\alpha_{t-\Delta}(1-\bar\alpha_{t})}{\bar\sigma^2_{t-\Delta}(\bar\alpha_{t-\Delta}-\bar\alpha_{t})}}_{1/\tilde \sigma_t^2 }(x^{t-\Delta})^2\\
&\quad\quad\ \ \ -2\underbrace{(\frac{\bar\alpha_{t}\bar\alpha_{t-\Delta}}{\bar\alpha_{t-\Delta}-\bar\alpha_{t}}x^t+\frac{\bar\alpha_{t-\Delta}}{1-\bar\alpha_{t-\Delta}}x^0)}_{\tilde\mu_t/\tilde \sigma_t^2}x^{t-\Delta}+C)).
\end{aligned}
}\color{black}\end{equation}

\lemma{For Gaussian distribution $\mathcal{N}(\mu,\sigma^2)$, its expansion is $G(x|\mu,\sigma^2)\propto \exp(-\frac{(x-\mu)^2}{2\sigma^2})=\exp(-\frac{1}{2}(\frac{x^2}{\sigma^2}-2\frac{\mu}{\sigma^2}x+C))$}.\rm

Then, we can express the variables $\tilde\mu_t$ and $\tilde\sigma^2_t$ as:
\begin{equation}\color{rbrown}{
\begin{aligned}
\tilde \sigma_t^2=\frac {\bar\sigma^2_{t-\Delta}(\bar\alpha_{t-\Delta}-\bar\alpha_{t})}{\bar\alpha_{t-\Delta}(1-\bar\alpha_{t})},
\tilde\mu_t=\frac{\bar\alpha_{t}(1-\bar\alpha_{t-\Delta})x^t+(\bar\alpha_{t-\Delta}-\bar\alpha_{t})x^0}{\bar\alpha_{t-\Delta}(1-\bar\alpha_{t})}.
\end{aligned}
}\color{black}\end{equation}

We note that in Song et al.~\cite{song2020improved}, a variable variance $\Sigma_\eta$ is introduced, and then the conditional distribution $p_\theta(x^{t-\Delta}|x^t)$ can be conditionally updated as an implicit process without random noise injection. So we further define:
\begin{equation}\color{black}{
    p_\theta(x^{t-\Delta}|x^t)=p_\theta(x^{t-\Delta}|x^t,x^0)= \mathcal{N}(x^{t-\Delta};k\ x^0+m\ x^t,\Sigma_\eta{\rm I}),
}\color{black}\end{equation}
where we use undetermined coefficients $k$ and $m$ to represent the variable $\mu_\theta$, and define $\Sigma_\eta=\eta\Sigma=\eta\tilde \sigma_t^2$. In the subsequent section, we derive the expression $\mu_\theta$ controlled by the coefficient $\Sigma_\eta$.

First, replace variable $x^t$ by $q(x^t|x^0)$:
\begin{equation}\color{black}{
    p_\theta(x^{t-\Delta}|\bcancel{x^t},x^0)= \mathcal{N}\left(x^{t-\Delta};(k+m)x^0,\left(\frac{1-\bar\alpha_t}{\bar\alpha_t}m^2+\Sigma_\eta\right){\rm I}\right).
}\color{black}\end{equation}
Since there is an equivalence between $p_\theta(x^{t-\Delta}|\bcancel{x^t},x^0)$ and $q(x^{t-\Delta}|x^0)$, we have:
\begin{equation}\color{rbrown}{
    m=\sqrt{\frac{(1-\bar\alpha_{t-\Delta}(1+\Sigma_\eta))\bar\alpha_{t}}{\bar\alpha_{t-\Delta}(1-\bar\alpha_{t})}}}\color{black}\quad  {\rm and}\quad k=1-m,
\end{equation}
and we have the relationship $\mu_\theta =(1-m)x^0+m\ x^t$,
which can be reparametrized into a form:
\begin{equation}\color{rbrown}{
    \mu_\theta =x^t+(\sqrt{\frac{1-\bar\alpha_{t-\Delta}(1+\Sigma_\eta)}{\bar\alpha_{t-\Delta}}}-\sqrt{\frac{1-\bar\alpha_{t}}{\bar\alpha_{t}}})z_\theta(x^t).
}\color{black}\end{equation}

Previous Displacement and gradient-based works~\cite{zhang2020pointfilter,luo2021score,song2021scorebased} have found that modeling gradient gives better results than modeling noise.
\lemma{For Gaussian variable $x\sim p(x)= \mathcal N(x;\ \mu,\Sigma)$, we can estimate the mean $\mu$ using $x$ and the gradient of log-likelihood $\log{p(x)}$:
\begin{equation}\label{}
    \mu=x+\Sigma\nabla_x\log{p(x)}.
\end{equation}}\rm
Plug Lemma.2 into $q(x^t|x^0)$, we further define the score as the gradient $-\bar{\sigma}_t z_\theta\propto\bar{\sigma}^2_t\nabla_x\log{[p_{\theta}(x^t|x^0)]}=s_\theta(x^t)$. Then $\mu_\theta$ can be expressed as:
\begin{equation}\color{rbrown}{
    \mu_\theta  =x^t+(1-\sqrt{\frac {(1-\bar\alpha_{t-\Delta}(1+\Sigma_\eta))\bar\alpha_t}{(1-\bar\alpha_t)\bar\alpha_{t-\Delta}}})s_\theta(x^t).
}\color{black}\end{equation}

\section{Derivation of the loss function}

\begin{figure*}[h]
    \centering
    \includegraphics[width=\linewidth]{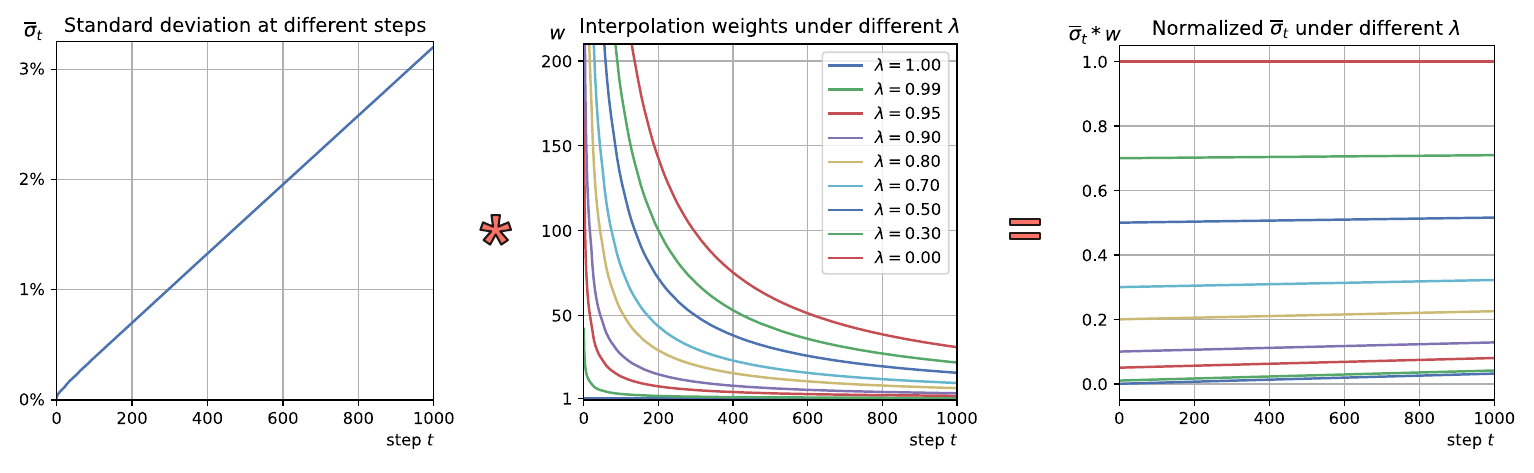}
    \caption{\textbf{Visual of standard deviation $\bar{\sigma}_t$ at different steps under different $\lambda$} The left plot visualizes the noise scale in the diffusion process which is represented by the parameter $[T,\beta_{T}]$. The middle visualizes the interpolated weights $w=(1-\lambda)/\bar{\sigma}_t+\lambda$ in loss $\mathcal L$. The right visualizes the product of previous two results, representing the normalized scale at different steps $t$ in loss $\mathcal L$.}
    \label{fig:supp_lw}
\end{figure*}

Following DDPM~\cite{ho2020denoising}, our trainable objective $\mathcal{L}$ is obtained by deriving the variational lower bound on negative log-likelihood:
\begin{equation}
\begin{aligned}
\mathbb E[&-\log p_\theta(x^0)]\\
&\le \mathbb E_q\bigg[-\log \frac {p_\theta(x^{0:{T}})}{q(x^{1:{T}}|x^0)}\bigg]\\
&=E_q\bigg[-\log \frac{p(x^{T})}{q(x^{T}|x^0)}-\sum_{t> 1}\log \frac {p_\theta(x^{t-1}|x^{t})}{q(x^{t-1}|x^t,x^0)}-{\log {p_\theta(x^{0}|x^{1})}}\bigg]\\
&={\color{rbrown}E_q\bigg[\sum_{t> 1,\Delta}D_{KL}({q(x^{t-\Delta}|x^t,x^0)} \ ||\ {p_\theta(x^{t-\Delta}|x^{t})})\bigg]
}\\
&\quad\quad\ +D_{KL}({q(x^{T}|x^0)}\ ||\ {p(x^{T})})-\log {p_\theta(x^{0}|x^{1})}\\
&\le{\color{rbrown} E_q\bigg[\sum_{t> 1,\Delta}D_{KL}({q(x^{t-\Delta}|x^t,x^0)} \ ||\ {p_\theta(x^{t-\Delta}|x^{t})})\bigg]
}\\
&=:\mathcal{L}(x^{0:{T}},\{\beta_t\}^{T}_{t=1}).
\end{aligned}
\end{equation}
\color{rbrown}{where $\sum_{t> 1,\Delta}$ denotes a summation with $t$ taking values in steps $\Delta$.}\color{black}

\lemma{For multivariate Gaussian distributions $\mathcal{N}_0(\mu_0,\Sigma_0)$ and $\mathcal{N}_1(\mu_1,\Sigma_1)$, their KL divergence can be expressed as 
\begin{equation}
\begin{aligned}
D_{KL}(\mathcal N_0||\mathcal N_1)&=\\
\frac 1 2\Big({\rm tr}{\Sigma_1^{-1}\Sigma_0}&-k+(\mu_1-\mu_0)^T\Sigma_1^{-1}(\mu_1-\mu_0)+\ln\frac{\det \Sigma_1}{\det \Sigma_0}\Big).
\end{aligned}
\end{equation}}\rm

After simplifying the summation term, if we continue to simplify $\mathcal L$, we need to expand the KL divergence term:
\begin{equation}
\begin{aligned}
    D_{KL}&({\color{rbrown}{q(x^{t-\Delta}|x^t,x^0)} \ ||\ {p_\theta(x^{t-\Delta}|x^{t})}})\\
    &=\frac 1 2(3-3+{\Sigma}^{-1}\vert\vert\tilde\mu_t-\mu_\theta\vert\vert^2+\log 1)\\
    &\le \vert\vert s_\theta(x^t)-S(x^t) \vert\vert^2,
\end{aligned}
\end{equation}
where, $-\bar{\sigma}_t z\propto \bar{\sigma}_t^2\nabla_x\log{[q(x^t|x^0)]}=S(x^t)$. Here, we obtain the first loss function, which is computed directly at the score:
\begin{equation}
    \mathcal{L}_1=\mathbb E_{q,t>0}\big[\big\vert\big\vert s_\theta(x^t)-S(x^t)\big\vert\big\vert^2_2\big].
\end{equation}
It is evident that $\mathcal L_1$ is more concerned about large-scale noise, e.g. the gradient with $\bar\sigma=3\%$ is triple larger than the one with $\bar\sigma=1\%$.

Notably, we can observe similar relationships between the score and Gaussian distribution in $-\bar{\sigma}_t z_\theta\propto s_\theta(x^t)$ and $-\bar{\sigma}_t z\propto S(x^t)$. Then, we propose the loss with normalized standard deviation:
\begin{equation}
    \mathcal{L}_2=\mathbb E_{q,t>0}\big[\big\vert\big\vert \frac{1}{\bar{\sigma}_t}(s_\theta(x^t)-S(x^t))\big\vert\big\vert^2_2\big].
\end{equation}
Here, this weight $1/\bar{\sigma}_t$ is very large for extremely small noise, e.g. $\bar{\sigma}_t=0.1\%$.
In our experiments, we observed that $\mathcal L_2$ is slightly better at learning the score of point clouds with small-scale noise compared to $\mathcal L_1$.
We hope our model $s_\theta$ to have a balanced performance on both small and large noise, so we chose the interpolated weight in our implementation:
\begin{equation}\label{loss}
    \mathcal{L}=\mathbb E_{q,t>0}\bigg[\Big\vert\Big\vert \Big(\frac{1-\lambda}{\bar{\sigma}_t}+\lambda\Big)(s_\theta(x^t)-S(x^t))\Big\vert\Big\vert^2_2\bigg].
\end{equation}
It is evident that the loss $\mathcal L$ also is the variational bound of $\mathbb E[-\log p_\theta(x^0)]$. In our experiments, we set $\lambda=0.99$ because $1/\bar{\sigma}_t$ is much larger than 1.

The left picture of Fig.~\ref{fig:supp_lw} plots the standard deviation of the coupled noise in the point clouds, where its variation in our diffusion process under $[{T}=1000,\beta_{T}=2e^{-6}]$ is approximately linear.
When $\lambda=0$, the scores in loss are normalized to the standard scale. This means the scores of small-scale noisy point cloud, which are easily ignored under $\lambda=1$, become more significant than those of large noise.
However, in our experiment, we found that the $s_\theta$ obtained by $\lambda=0$ is more likely to generate bumpy details when denoising large-scale noisy point clouds.
In contrast, the $s_\theta$ obtained by $\lambda=1$ tends to smooth the details when denoising small-scale noisy point clouds.
For more general performance, we select $\lambda$ between 0 and 1.
Further, since the weight $1/\bar{\sigma}_t$ is obviously larger than 1, we set $\lambda=0.99$ in our experiment.

{\color{rblue}
\section{Further Ablations on Iterative Strategy }

\begin{table*}[t]
    \centering
    \caption{\textbf{Ablation study of the Iterative Strategy and Adaptive Schedule Arrangement.}}
    \label{tab:ab_is}
    \resizebox{\linewidth}{!}{
    \renewcommand\arraystretch{1}
    \begin{tabular}{lccccccccccc|ccccccccccc}
    \toprule
           & \multicolumn{11}{c}{CD ($10^4$)} & \multicolumn{11}{|c}{P2M ($10^5$)} \\
    PU-Set & \multicolumn{3}{c}{1\%}& & \multicolumn{3}{c}{2\%}& & \multicolumn{3}{c}{3\%} & \multicolumn{3}{|c}{1\%}& & \multicolumn{3}{c}{2\%}& & \multicolumn{3}{c}{3\%}\\
    \cmidrule{2-4}\cmidrule{6-8}\cmidrule{10-12}\cmidrule{13-15}\cmidrule{17-19}\cmidrule{21-23}
     Ablation & 10K    & 50K    & 100K   & & 10K    & 50K    & 100K   & & 10K    & 50K    & 100K   & 10K    & 50K    & 100K   & & 10K    & 50K    & 100K   & & 10K    & 50K    & 100K\\  
    \midrule          
        SBD
        & 2.52   & 0.71   & 0.42   & & 3.68   & 1.29   & \firc{1.04}   & & 4.71   & 1.93   & 1.75   & 4.63   & 1.50   & 1.12   & & 10.7   & 5.66   & \firc{5.78}   & & 19.4   & 10.4   & 10.8   \\   
        SBD (+our)
        & \firc{2.41}   & \firc{0.70}   & \firc{0.41}   & & \firc{3.61}   & \firc{1.20}   & 1.34   & & -      & -      & -      & \firc{3.95}   & \firc{1.41}   & \firc{1.05}   & & \firc{9.97}   & \firc{5.37}   & 7.58   & & -   & -   & -   \\ 
    \midrule
        PSR
        & 2.42   & 0.65   & 0.37   & & 3.44   & 1.03   & 0.62   & & 4.17   & 1.40   & 1.06   & 3.31   & 0.75   & \firc{0.49}   & & 8.17   & 3.30   & 2.11   & & 13.4   & 5.82   & 4.93   \\ 
        PSR (+our)
        & \firc{2.33}   & \firc{0.64}   & \firc{0.36}   & & \firc{3.32}   & \firc{0.99}   & \firc{0.59}   & & \firc{4.03}   & \firc{1.31}   & \firc{1.02}   & \firc{3.02}   & \firc{0.69}   & 0.52   & & \firc{7.42}   & \firc{3.05}   & \firc{2.02}   & & \firc{12.3}   & \firc{5.22}   & \firc{4.82}   \\ 
    \midrule
        w/o align
        & \firc{2.09}   & \firc{0.59}   & \firc{0.30}   & & \firc{2.99}   & \firc{0.78}   & \firc{0.50}   & & {3.63}   & {1.19}   & {1.11}   & \firc{2.03}   & {0.52}   & \firc{0.29}   & & \firc{5.07}   & {1.69}   & {1.46}   & & {9.44}   & {4.51}   & {5.61}   \\ 
        Full
        & \firc{2.09}   & \firc{0.59}   & \firc{0.30}   & & \firc{2.99}   & \firc{0.78}   & \firc{0.50}   & & \firc{3.62}   & \firc{1.18}   & \firc{1.09}   & {2.04}   & \firc{0.51}   & \firc{0.29}   & & {5.09}   & \firc{1.68}   & \firc{1.44}   & & \firc{9.41}   & \firc{4.45}   & \firc{5.51}   \\  
    \bottomrule
    \end{tabular}
}
\end{table*}

We conduct a further ablation study to validate our iterative strategy by directly replacing the iterative algorithms of SBD and PSR, denoted as $+our$ in Table~\ref{tab:ab_is}. Since SBD does not fully adhere to its iterative algorithm when denoising 3\% of inputs, we only compare it on the 1\% and 2\% portions of the PUNet test set. Consequently, the iterative strategy in our method effectively avoids surface collapse caused by over-iteration, particularly for sparse point clouds.
}

{\color{rblue}
\section{More Discussion on Schedule Arrangement }
\RE{
During training (Section~\ref{alg:training}), our $\Delta$ sampling strategy is motivated by the formulation presented in Section~\ref{sec:diffusion}, where the diffusion process and sampling scheme are based on a Markov chain defined by training schedule $\{\beta_t\}$. Consequently, we design the uniform $\Delta$ sampling strategy to accurately model the reverse recovery sequence described in Eq.~\ref{dif:pThetaDefine}, ensuring that our denoising network satisfies both Section~\ref{sec:diffusion} and the training schedule.
}

In Section~\ref{sec:algorithm}, we employ an adaptive schedule aligned to the training schedule during inference.
Table~\ref{tab:ab_is} presents an ablation study that validates the effectiveness of aligning the schedules used during training and inference.
We replace the search for the corresponding timestep $\hat\tau\in\{0,\dots,T\}$ as directly searching for $\beta_{\hat\tau}\in R^+$, denoted as $w/o\ align$:
\begin{equation}\label{eq:aboptimalbeta}
    \beta_{\hat \tau}=\mathop{\rm argmin}_{\beta_{\hat \tau}\in R^+}\ \big\vert \bar\sigma^2_{\hat \tau}-\bar\sigma^2 \big\vert.
\end{equation}
The search terminates when $\big\vert \bar\sigma^2_{\hat \tau}-\bar\sigma^2 \big\vert<10^{-5}$.
Then, we can obtain $\{\beta_t\}^{\hat\tau}_{t=1}$ with linear interpolation (assuming $\beta_0=0$) irrespective of the alignment operation.

In contrast, the full implementation of our method uses the adaptive schedule aligned to the training schedule. This more straightforward manner provides slightly more robust performance under varying levels of noise.
}

\end{document}